\documentclass[10pt,twocolumn,letterpaper]{article}

\usepackage{iccv}
\usepackage{times}
\usepackage{epsfig}
\usepackage{graphicx}
\usepackage{amsmath}
\usepackage{amssymb}
\usepackage{cite}
\usepackage{booktabs}
\usepackage{xcolor,colortbl}
\usepackage{multirow}
\usepackage{makecell}
\usepackage{adjustbox}
\usepackage{subfigure}

\usepackage{booktabs}
\usepackage{multirow}
\usepackage{makecell}

\definecolor{shadecolor}{rgb}{0.92,0.92,0.92}

\makeatletter %
\@namedef{ver@everyshi.sty}{}
\makeatother
\usepackage{tikz}

\newcommand{\name}{0}
\newcommand{\h}{0}
\newcommand{\w}{0.15}
\newcommand{\wa}{0.15}
\newlength \g

\usepackage[pagebackref=true,breaklinks=true,letterpaper=true,colorlinks,bookmarks=false]{hyperref}

 \iccvfinalcopy % *** Uncomment this line for the final submission

\def\iccvPaperID{6340} % *** Enter the ICCV Paper ID here

\ificcvfinal\fi

\begin{document}

%%%%%%%%% TITLE
\title{Inheriting Bayer's Legacy: Joint Remosaicing and Denoising for Quad Bayer Image Sensor}

\author{$\text{Haijin Zeng}^1$, $\text{Kai Feng}^2$, $\text{Jiezhang Cao}^4$, $\text{Shaoguang Huang}^3$, $\text{Yongqiang Zhao}^2$, \\ $\text{Hiep Luong}^1$, $\text{Jan Aelterman}^1$, and 
	$\text{Wilfried Philips}^1$\\
	$^1\text{IMEC-IPI-UGent}$, $^2\text{NWPU}$, $^3\text{CUG}$, $^4\text{ETH Zurich}$ \\
	
	{\tt\small haijin.zeng@ugent.be}
}

\maketitle
\ificcvfinal\thispagestyle{empty}\fi

%%%%%%%%% ABSTRACT
\begin{abstract}

Pixel binning based Quad sensors have emerged as a promising solution to overcome the hardware limitations of compact cameras in low-light imaging. However, binning results in lower spatial resolution and non-Bayer CFA artifacts. To address these challenges, we propose a dual-head joint remosaicing and denoising network (DJRD), which enables the conversion of noisy Quad Bayer and standard noise-free Bayer pattern without any resolution loss. DJRD includes a newly designed Quad Bayer remosaicing (QB-Re) block, integrated denoising modules based on Swin-transformer and multi-scale wavelet transform. 
The QB-Re block constructs the convolution kernel based on the CFA pattern to achieve a periodic color distribution in the perceptual field, which is used to extract exact spectral information and reduce color misalignment. 
The integrated Swin-Transformer and multi-scale wavelet transform capture non-local dependencies, frequency and location information to effectively reduce practical noise. 
By identifying challenging patches utilizing Moiré and zipper detection metrics, we enable our model to concentrate on difficult patches during the post-training phase, which enhances the model's performance in hard cases. Our proposed model outperforms competing models by approximately 3dB, without additional complexity in hardware or software.

%-------------
\end{abstract}

%%%%%%%%% BODY TEXT
\section{Introduction}

In recent years, smartphones have emerged as the most popular choice for photography. Nevertheless, due to the demand for portable devices, smartphones are designed with compact and cost-efficient cameras, which pose a challenge in capturing high-quality images comparable to those produced by DSLR cameras \cite{ignatov2017dslr}. 
\begin{figure}[t!]
\centering
\includegraphics[width=0.48\textwidth]{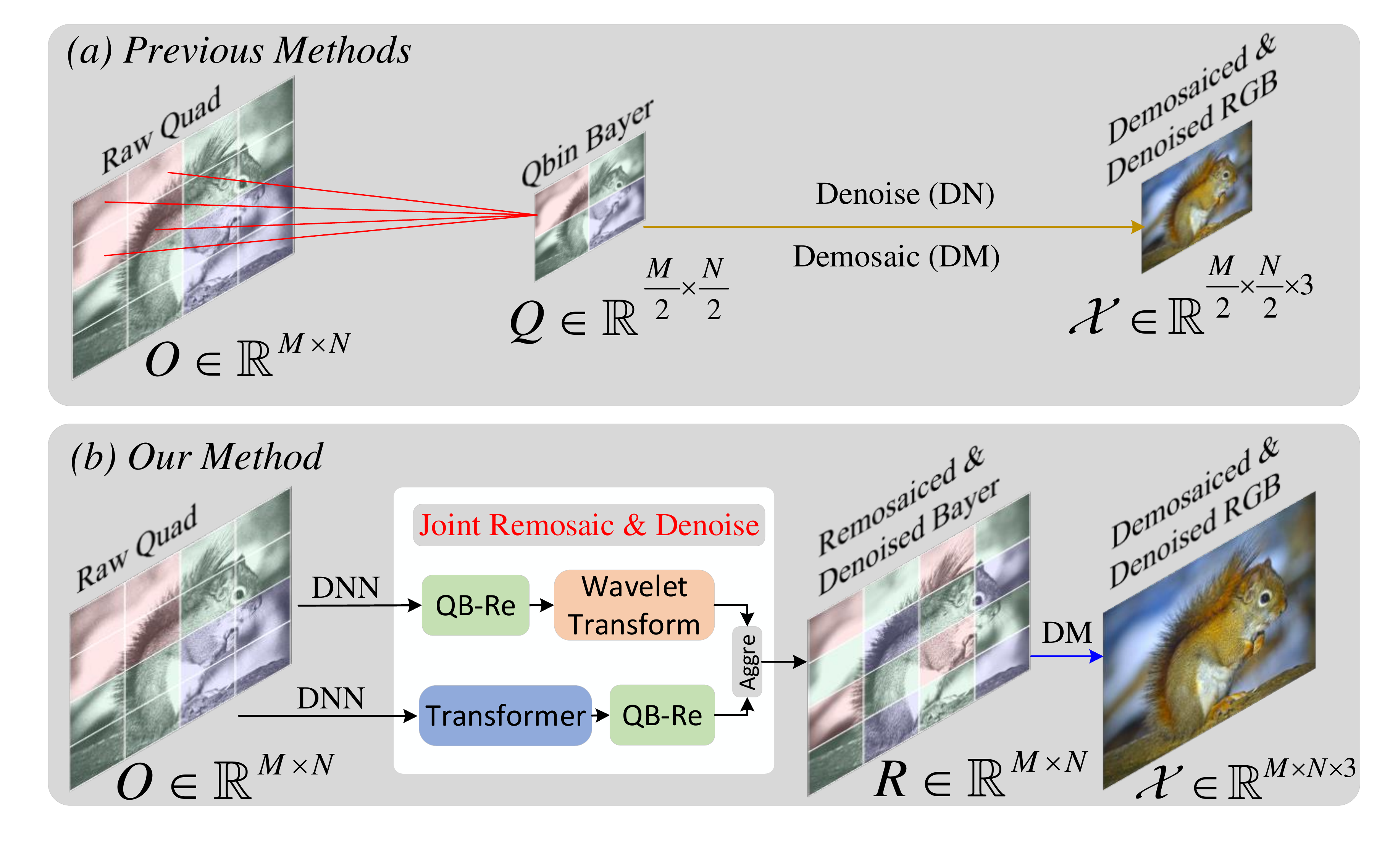}
\vspace{-8mm}
\caption{To address the raw Quad Bayer demosaicing problem, previous methods mainly focus on models demosaic the Quad directly or average the neighbor $2 \times 2$ pixels to one,  while we promote a novel view by designing a joint but flexible remosaic and denoise module to convert noisy Quad Bayer to full-resolution clean Bayer. It allows any advance in Bayer CFA tools to be directly applicable in Quad Bayer.} 
\label{fig:compare_Qbin_remosaic}
\end{figure}

% \begin{figure*}[t!]
%     \centering
%     \includegraphics[width=1.0\textwidth]{images/CVPR_comparison_5.jpg}
%     \caption{Example of Joint demosaicing and denoising on Quad Bayer CFA, our model consists of the proposed joint remosaic and denoise network and a simple camera ISP \cite{yang2022mipi}.}
%     \label{fig:example_show}
% \end{figure*}

Pixel binning using Quad Bayer Color Filter Array (CFA) technology has been recognized as a promising approach for producing high-quality images under low-light conditions, as evidenced in previous studies \cite{yoo2015low,kim2019high}. The Quad Bayer CFA pattern is composed of periodic $2 \times 2$ cells that are designed to capture two consecutive homogenous pixels of the same color in two spatial dimensions. By averaging four pixels within a $2 \times 2$ neighborhood, the Quad Bayer CFA can capture larger pixels and collect twice as much light intensity as the standard Bayer pattern, resulting in high-sensitivity and high-resolution imaging with low energy consumption.

Apart from improving image quality in low-light conditions, the Quad Bayer CFA also enables original equipment manufacturers (OEMs) to create higher-resolution sensors for mobile photographers \cite{kim2020deep,BJDD}. This feature makes it possible to produce high-resolution videos, such as 8K videos, which allow for high-definition imaging even when using digital zoom on smartphones. Therefore, the Quad Bayer CFA is commonly used in smartphone cameras. For example, leading mobile companies have utilized Quad Bayer CFA in conjunction with 108-megapixel image sensors in their latest flagship smartphones, e.g., iPhone 14 Pro, providing a versatile photography experience for enthusiastic mobile photographers \cite{SAGAN}, as shown in Fig.~\ref{fig:galaxy}.

Reconstructing RGB image from raw Quad Bayer mosaic can be achieved by averaging the $2 \times 2$ neighboring pixels and applying a demosaicing algorithm designed for the Bayer CFA. However, this approach involves downsampling the Quad Bayer to Bayer, leading to a quarter of the original image resolution, as depicted in Fig.~\ref{fig:compare_Qbin_remosaic}. An alternative approach is Quad Bayer demosaicing, which aims to directly generate RGB images from raw Quad Bayer data, as demonstrated in previous works \cite{BJDD,SAGAN}. Nevertheless, demosaicing on Quad Bayer data often produces visual artifacts due to the six color components of the Quad Bayer CFA being located differently than the three color components of the standard Bayer CFA \cite{kim2021recent}. This also implies that Quad Bayer CFA is more susceptible to aliasing compared to Bayer CFA during demosaicing.
In addition, even with a reliable Quad Bayer demosaicing algorithm, it is necessary to redesign the current sophisticated Bayer image signal processor (ISP) for Quad sensor with new arrangement of color components in both software and hardware. 
Against the above issues, we propose a dual-head joint remosaicing and denoising network, which enables conversion of noisy Quad Bayer to a standard clean Bayer mosaic without any loss in resolution. It facilitates the use of all the software and hardware designed for classic Bayer CFA, and allows any advance in Bayer CFA tools to be directly applicable in our approach. Therefore, the impact is far-reaching, extending beyond just remosaicing.

Firstly, we propose a novel and efficient basic component, the Quad Bayer remosaicing (QB-Re) block, which utilizes Quad Bayer CFA guided convolution to extract spectral information and reduce color misalignment. This design constructs the convolution kernel based on the CFA pattern, with the same weights assigned to pixels in the same relative positions within the CFA, and periodic weight changes as the kernel slides. This results in a periodic color distribution in the perceptual field, ensuring that neighboring pixels with the same color have similar spectral distributions. Additionally, we introduce a Quad Bayer CFA pooling layer that refines features with the same relative CFA position, instead of using common pooling methods.

Secondly, based on the proposed QB-Re block, we present a dual-head joint remosaicing and denoising network, named DJRD. It leverages the Swin-Transformer and multi-scale wavelet transform to model non-local dependencies, while simultaneously capturing frequency and location information of feature maps with limited computation.
% Thirdly, to enhance the model's performance on hard cases, in post-training phase, we fine-tune DJRD on difficult patches selected by using hard patches detection metrics. Such a fine-tune strategy make DJRD more compatible to practical scenarios.
Thirdly, to make the DJRD model more robust and better suited for practical scenarios, in the post-training phase, we fine-tune our DJRD on difficult image patches. These patches were selected using hard patch detection metrics, which helped identify regions where the model was struggling to make accurate predictions. 
Overall, our contributions are four-fold:
\begin{itemize}
    \item We propose DJRD, a novel dual-head joint remosaicing and denoising network to reconstruct clean classic Bayer images from noisy Quad Bayer mosaic without any resolution loss.

    \item We present a Quad Bayer CFA-driven CNN architecture to exploit the spatial-channel correlation of Quad Bayer.

    \item We enhance DJRD's capability in challenging scenarios through fine-tuning it on difficult cases by using hard patches finding metrics.

    \item Extensive experiments show that DJRD establishes new state-of-the-arts on various datasets for joint Quad Bayer remosaicing and denoising task.

\end{itemize}
% Then, we generate RAW Quad-Bayer in various scenes with a fixed camera setting and augment them into RAW images with random rotations steps, and random left-right mirror images.
% This augments the training data provides some rotational and translational invariance. 

% Our contributions are listed as follows:

% \begin{itemize}
%     \item A Quad CFA guided attention remosaic network is proposed as the first attempt to convert the Quad to full resolution Bayer.

%     \item A flexible joint remosaic and denoise deep model is proposed. 
%     For the question of how to combine denoising and remosaicing to reconstruct full Bayer images, we propose a parallel pipeline, which integrates both the advantages of denoising first and remosaicing first.
    
%     \item With the designed parallel metric, swin-conv-attention and multi-level wavelet-CNN are integrated into the proposed network, to adaptively cope with cases that noise shows variable strength, which can significantly improve the practicability for real images.
    
%     \item To enhance the mode practicability in challenging images prone to moiré and artifacts, we introduce and enhance reflectional and rotational invariance through the network design.
%     \item We also create a new dataset with paired Quad, Bayer and RGB images, together with a better training set with additional hard cases by inverting the camera ISP and employing metrics to identify difficult patches and techniques for mining community photographs for such patches.
% \end{itemize}

 

\section{Related Work}
% Here, some related works of Bayer demosaicing, joint demosaicing and denoising, and Non-Bayer demosaicing are reviewed briefly.

\subsection{ Classic Bayer Demosaicing}

Bayer demosaicing is a low-level image signal processing (ISP) task that has been extensively researched for several decades. The main objective of demosaicing is to reconstruct RGB images from observed mosaic images captured by a sensor with a Bayer filter. Interpolation-based methods are commonly employed to demosaic R, G, and B channels by using various linear or non-linear interpolation techniques, such as bilinear interpolation \cite{Demosaic_bilinear}, directional linear \cite{Demosaic_linear_minimum}, and others. Although these methods are spatially invariant and effective for a single color channel, they can produce pseudocolor at joints with different color variations.
To overcome this color issue, several demosaicing methods have been developed, such as the edge-adaptive algorithm \cite{demosaic_edge}, reconstruction-based models \cite{demosaic_markov}, and frequency domain filtering  \cite{demosaic_filter,demosaic_frequency}. However, these conventional models still have limitations, such as visually disturbing artifacts like moiré patterns appearing on challenging high-pass regions when enlarging the local patches  \cite{demosaic_selfguidance}.
Recently, deep learning-based methods have been proposed and shown superior performance on various image processing tasks, including demosaicing, e.g.,  \cite{demosaic_selfguidance,demosaic_deep_full,demosaic_resnet,feng2021mosaic,stojkovic2019effect,zhang2022deep,DM_searching,mei2019higher,ahmed2023image,dong2022abandoning}. These methods utilize deep neural networks to learn a mapping between observed mosaic images and their corresponding RGB images, achieving promising performance.

\subsection{Joint Demosaicing and Denoising}

% Bayer demosaicing has been extensively studied as mentioned. 
% However, in the imaging process of real scenes, the observed raw images are often degraded by various noise processes due to hardware accuracy, imaging environment and other limitations \cite{demosaic_selfguidance,BJDD}. 
% Therefore, algorithms designed only for demosaicing cannot be directly applied to real scenes. 
% To address this problem, some approaches based on hybrid solution frameworks for image processing have been proposed, which focus on performing denoising and demosaicing simultaneously \cite{demosaic_joint2,demosaic_joint1, ehret2019joint,dewil2023video,guo2021joint}.
% Such joint denoising and demosaicing strategies additionally take into account more realistic noise factors in the imaging process and reduce to some extent the error accumulation caused by the distributed execution of each ISP. 
% As a result, they tend to achieve relatively good performance on real data, compared to models that process mosaics independently.

Bayer demosaicing is a widely researched topic in the field of imaging. However, real-world raw images are often corrupted by various types of noise due to hardware limitations and environmental factors \cite{demosaic_selfguidance,BJDD,liu2019learning}, among others. Therefore, demosaicing algorithms that are solely designed for this task cannot be directly applied to real scenes. To overcome this issue, hybrid solution frameworks for image processing have been proposed that simultaneously address both denoising and demosaicing \cite{demosaic_joint2,demosaic_joint1, ehret2019joint,dewil2023video,guo2021joint}. By considering more realistic noise factors in the imaging process, these joint denoising and demosaicing methods reduce error accumulation caused by the distributed execution of each image signal processor (ISP) and thus achieve relatively better performance on real data compared to models that process mosaics independently.

\subsection{Non-Bayer Demosaicing}

% Although deep learning methods are already being applied for Bayer demosaicing from 2016, Quad Bayer has only been applied to cell phone cameras in the last few years, and there are only a few traditional or deep learning work on dedicated Quad Bayer demosaicing. 
% Two recent works focused on Quad Bayer demosaicing are PIPNet \cite{BJDD} and SAGAN \cite{SAGAN}, based on depth-spatial feature attention and adversarial spatial-asymmetric attention, respectively.
% These methods all perform demosaicing directly on Quad Bayer raw images. 
% Missing pixels are reconstructed by a black deep network that exploits intra-channel and inter-channel correlations in the raw image.

In recent years, deep learning techniques have been employed for demosaicing of Bayer pattern images. However, the use of Quad Bayer technology in cellphone cameras is a relatively new development, and there have been few studies on dedicated Quad Bayer demosaicing, either using traditional or deep learning approaches.
Two recent works, namely PIPNet \cite{BJDD} and SAGAN \cite{SAGAN}, have focused on Quad Bayer demosaicing. These methods employ depth-spatial feature attention and adversarial spatial-asymmetric attention, respectively, to perform demosaicing directly on the Quad Bayer raw images. In both methods, missing pixels are reconstructed using a deep neural network that leverages intra-channel and inter-channel correlations in the raw image.

\begin{figure}[t!]
    \centering
    \includegraphics[width=0.42\textwidth]{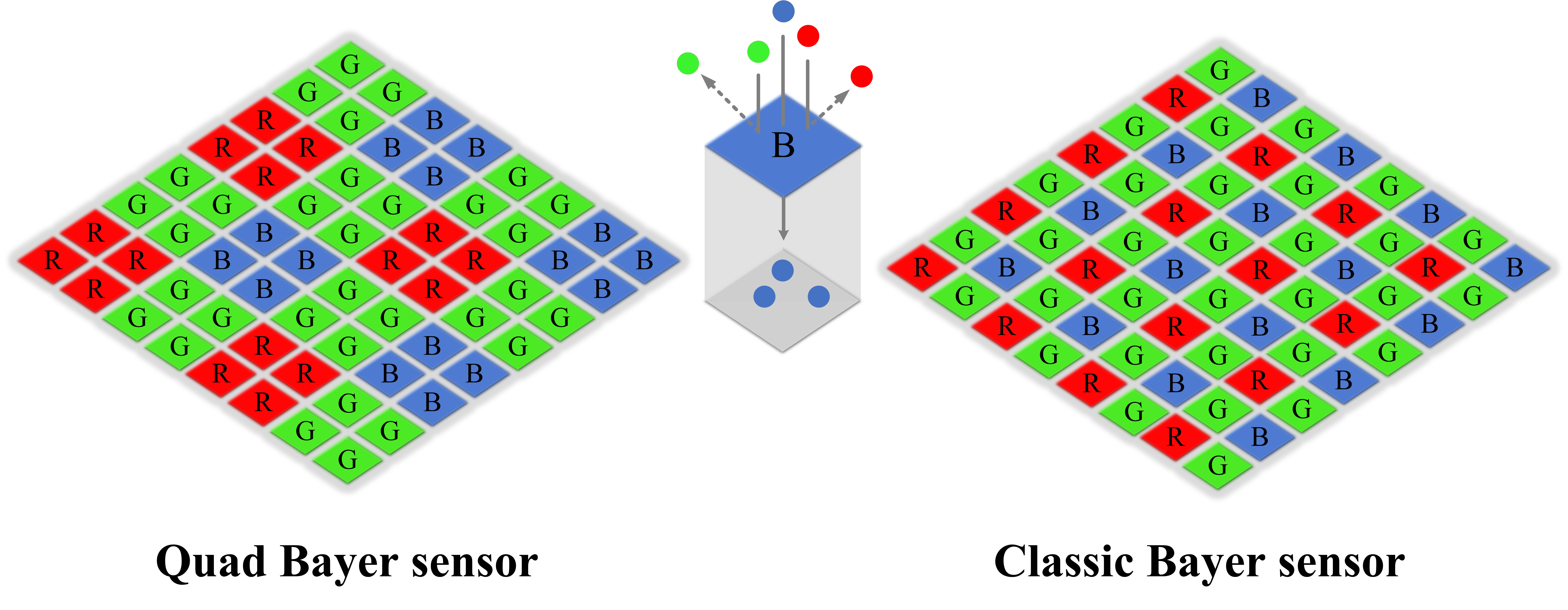}
    \includegraphics[width=0.44\textwidth]{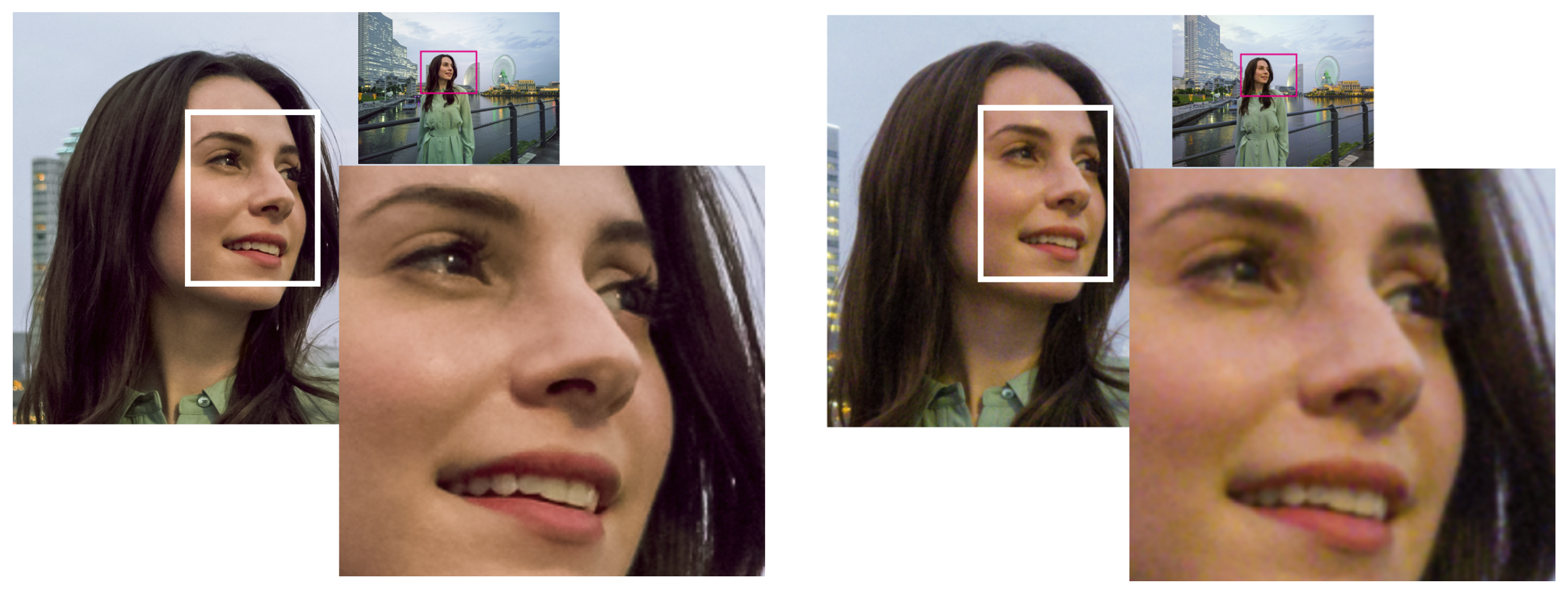}
    \vspace{-1mm}
    \caption{Quad Bayer and Bayer Color Filter Array (CFA) layouts, the pictures captured by image sensor with Bayer structure and image sensor with Quad Bayer structure (Sony IMX689). One can see that Quad Bayer prevents resolution loss in a low-illuminance environment and produces low-noise nightscape photo. Please zoom in for better view.} 
    \label{fig:galaxy}
\end{figure}

% \begin{figure}[t!]
%     \centering
%     \includegraphics[width=0.46\textwidth]{images/CVPR_Demo_quad_remosaic.jpg}
%     \caption{Visually disturbing artifacts in Quad Bayer demosaicing.}
%     \label{fig:artifacts_Quad_demosaic}
% \end{figure}

\begin{figure*}
\centering
\includegraphics[width=0.82\textwidth]{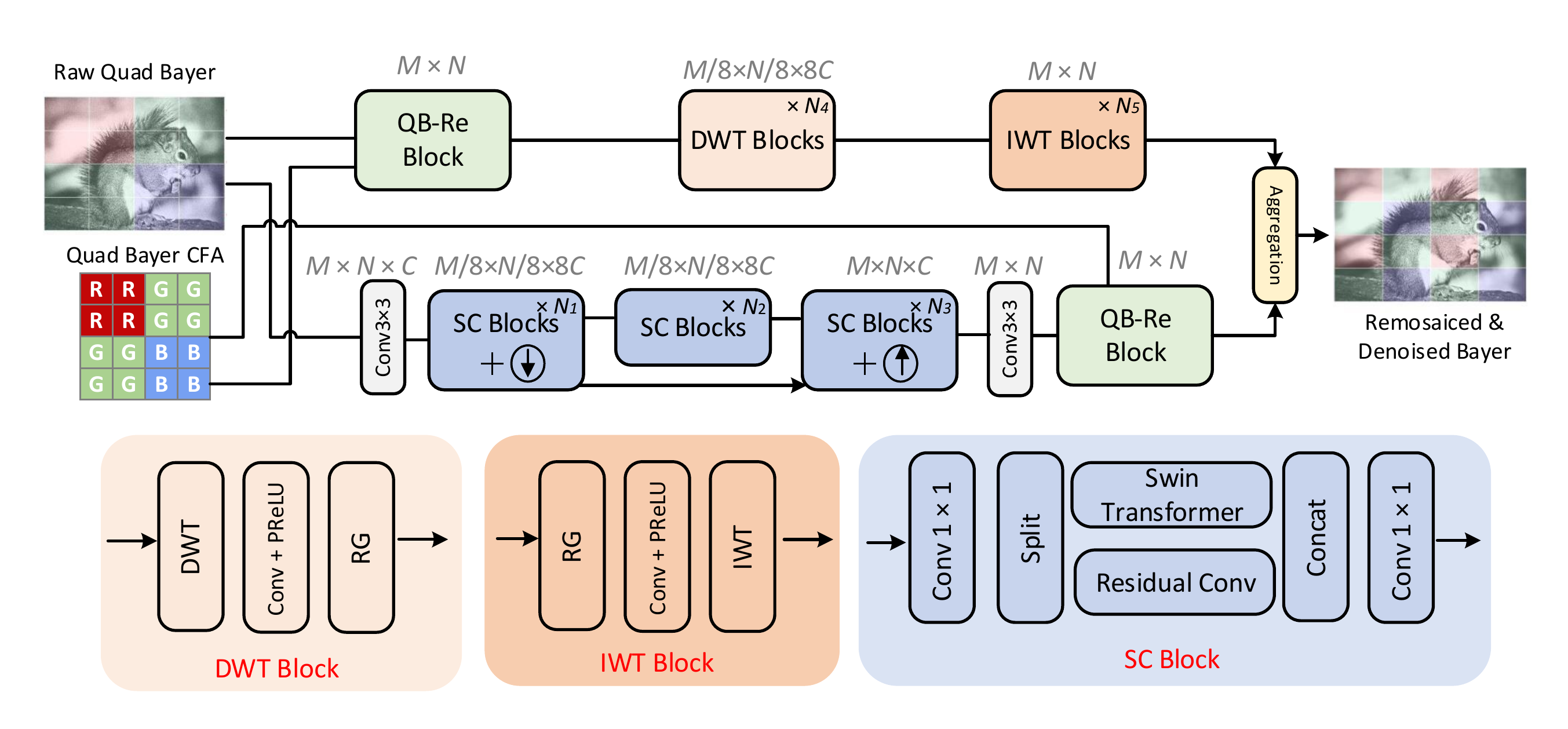}
\vspace{-4mm}
\caption{Overview of the proposed dual-head joint remosaicing and denoising network (DJRD) for Quad Bayer CFA based image sensor.}
\label{remosaic:sketch_U}
\end{figure*}

\section{Problem Formulation}

\emph{Quad Bayer Sensor VS. Classic Bayer}:
As shown in Fig. \ref{fig:galaxy}, the binning mode of the Quad sensor can produce superior image quality in low-light conditions when compared to the Bayer sensor, particularly in mobile devices such as smartphones, leading to its widespread use \cite{yang2022mipi, linear_demosaicing}. This subsection commences with an analysis of the disparities between the Quad Bayer and Bayer sensors utilizing the frequency structure matrix approach \cite{linear_demosaicing}. Additionally, we examine the advantages, limitations, and cost-effectiveness of these disparities, as well as the reason behind the Quad Bayer sensor's functionality in low-light environments.

% \begin{equation} \label{equ:Bayer_Quad}
% \textbf{CFA}_{\text {Bayer }}=\left[\begin{array}{ll}
% G & R \\
% B & G
% \end{array}\right],
% \textbf{CFA}_{\text{Quad}}=
% \left[\begin{array}{llll}
% G & G & R & R \\
% G & G & R & R \\
% B & B & G & G \\
% B & B & G & G
% \end{array}\right].
% \end{equation}
% where $R, G$ and $B$ denote the red, green and blue color filter, respectively. 

% \begin{equation}\label{eq:quad}
% \textbf{CFA}_{\text{Quad}}=
% \left[\begin{array}{llll}
% G & G & R & R \\
% G & G & R & R \\
% B & B & G & G \\
% B & B & G & G
% \end{array}\right].
% \end{equation}
Each $2 \times 2$ cell of Bayer has two green pixels, one red pixel and one blue pixel, while each $2 \times 2$ cell of Quad Bayer CFA consists of a single color as depicted in Fig.~\ref{fig:galaxy}.
To further identify the difference, we compare the Frequency Structure Matrices (FSMs) of these two CFAs, which represent the spectrum of image filtered with CFA \cite{linear_demosaicing}. 
Depicting the basic $4 \times 4 $ and $2 \times 2$ cell geometrical layout of Quad and Bayer sensor in Fig.~\ref{fig:galaxy} as matrices $\mathbf{C}_{\text{Quad}}$ and $\mathbf{C}_{\text{Bayer}}$, respectively, 
then, by using Discrete Fourier Transform (DFT) \cite{kim2021recent},
for Bayer matrix $\mathbf{C}_{\text {Bayer}}$, its FSM can be represented as follows:
\begin{equation}\label{equ:FSM_Bayer}
\mathbf{F}_{\text {Bayer }}= \operatorname{DFT} \left(\mathbf{C}_{\text {Bayer }}\right)=\left[\begin{array}{cc}
\mathbf{F}_L & 2\mathbf{F}_{c_2} \\
-2 \mathbf{F}_{c_2} & 2 \mathbf{F}_{c_1}
\end{array}\right],
\end{equation}
where $\mathbf{F}_L$ represents luminance component, $\mathbf{F}_{C_k}$ chrominance components, $k=1,2$, i.e., 
$\mathbf{F}_L=\frac{1}{4}(2 \mathbf{G+R+B}) $
$\mathbf{F}_{C 1}=\frac{1}{8}(2 \mathbf{G-R-B}) $
$\mathbf{F}_{C 2}=\frac{1}{8}(\mathbf{B-R})$.
% $$
% \begin{aligned}
% &F_L=\frac{1}{4}(2 G+R+B) \\
% &F_{C 1}=\frac{1}{8}(2 G-R-B) \\
% &F_{C 2}=\frac{1}{8}(B-R)
% \end{aligned}
% $$
Similarly, by applying DFT on Quad CFA matrix $C_{\text{Quad}}$, we have
\begin{equation}\label{equ:FSM}
    \mathbf{F}_{\text {Quad }}=\left[\begin{array}{cccc}
\mathbf{F}_L & \mathbf{F}_{c 2} & 0 & \mathbf{F}_{c 2} \\
-\mathbf{F}_{c 2} & 0 & 0 & \mathbf{F}_{c 1} \\
0 & 0 & 0 & 0 \\
-\mathbf{F}_{c 2} & -\mathbf{F}_{c 1} & 0 & 0
\end{array}\right].
\end{equation}
Equations (\ref{equ:FSM_Bayer}) and (\ref{equ:FSM}) demonstrate that Quad Bayer CFA possesses six color components located differently from the three color components of the standard Bayer CFA. This implies that Quad Bayer CFA is more susceptible to aliasing compared to the standard Bayer CFA, as previously noted \cite{linear_demosaicing}. However, increased aliasing can potentially result in better detail in shadows and highlights with promising subsequent demosaicing algorithms. While conventional demosaicing methods or averaging four pixels within a $2 \times 2$ cell can still be applied to Quad Bayer data, it may lead to severe visual artifacts \cite{kim2021recent} or resolution loss. In addition, reconstructing edges and details is a significant challenge for Quad Bayer sensors. These artifacts significantly reduce image quality, making them impractical for commercial ISPs. Therefore, more advanced methods are necessary to enhance image quality for Quad Bayer sensors.

% \subsection{Mathematical modelling of Problem}

% We substitute mosaic operator modelling Nona sensor (\ref{equ:Bayer_Quad}).
% $Y=\mathrm{M} \times \mathrm{X}+\mu$ where $\mathrm{Y} \in R^{12}$ denotes an observed raw image from sensor, $\mathrm{X} \in R^{\operatorname{9n}}$ is a reconstructed RGB image, $M$ represents a degradation matrix, $\mu \in$ $R^n$ - a noise vector, $\mathrm{n}$ - number of measurements.
% Matrix $\mathrm{M}$ here represents mosaic operator, where each color component is masked with corresponding mask, for instance $M_{e 1}^G$ in (\ref{equ:mask_example}).
% \begin{equation} \label{equ:mask_example}
% \begin{gathered}
% \mathrm{M}=\left[\begin{array}{lllll}
% M^R & M^0 & M^{\mathbb{E}}
% \end{array}\right] \\
% M_{9 i}^0=\left[\begin{array}{llll}
% 1 & 1 & 0 & 0 \\
% 1 & 1 & 0 & 0 \\
% 0 & 0 & 1 & 1 \\
% 0 & 0 & 1 & 1 
% \end{array}\right]
% \end{gathered}    
% \end{equation}

\section{Method}

In this section, we first describe the overall pipeline of our DJRD for joint remosaicing and denoising.
Then, we provide the details of the Quad Bayer remosaicing block. 
After that, we present the dual-head DJRD with integrated multi-scale wavelet transform and swin-transformer blocks. Then, we introduce the bottleneck data mining.

% In this section, we propose to convert Quad to Bayer first together with the parallel practical blind denoising, then demosaic the resulting Bayer with sophisticated ISP. 
% We then propose a joint remosaicing and denoising approach to Non-Bayer CFA sensor, named invariance guided parallel swin-conv-attention network for remosaic and denoise (ISRD). 
% In addition to our proposed model, we also create a dataset with paired Quad, Bayer and RGB images, along with a better training set with additional hard cases.

% \subsection{Overview}

\subsection{Overall Pipeline}

Fig. \ref{remosaic:sketch_U} shows the sketch of our DJRD, which is a dual-head network with three key blocks: our Quad Bayer remosaicing (QB-Re) block, Swin-Transform integrated residual convolution (SC) block \cite{SCUNet}, discrete wavelet transform (DWT) and inverse wavelet transform (IWT) blocks \cite{MWCNN}.
Specifically, given an observed Quad Bayer mosaic image $\mathbf{O} \in \mathbb{R}^{H \times W}$ with Pattern $\mathbf{Q} \in \mathbb{R}^{4 \times 4}$. Firstly, DJRD passes the $\mathbf{O}$ and $\mathbf{Q}$ through the proposed QB-Re block and a $3 \times 3$ convolution layer in parallel to extract feature map $\mathbf{X}_0, \mathbf{Y}_0$.
Next, on the one hand, 
$\mathbf{X}_0$ passes through $N_4$ DWT blocks, and following the convertibility of DWT, $\mathbf{X}_0$ is fed into $N_5$ IWT blocks, to up-sample low resolution feature maps, then the QB-Re block is used to reconstruct first primary Bayer output $\mathbf{X}_{out} \in \mathbb{R}^{M \times N}$.
On the other hand, 
$\mathbf{Y}_0$ are passed through $N_1$ Swin-Conv (SC) Blocks with down-sampling, $N_2$ Swin-Conv (SC) Blocks, and $N_1$ SC Blocks with up-sampling, $\mathbf{Y}_{l}$ is generated by a $3 \times 3$ convolution layer, and then be fed into QB-Re block to form the second primary Bayer output $\mathbf{Y}_{out} \in \mathbb{R}^{M \times N}$. 
Subsequently, the remosiced and denoised Bayer mosaic $\hat{\mathbf{I}}$ is obtained by aggregating the primary outputs: $\mathbf{X}_{out}$ and $\mathbf{Y}_{out}$.

In DJRD, QB-Re block is focusing on converting Quad Bayer pattern to Bayer pattern, by implementing CFA-driven convolution.
While, the SC block is used to model the non-local and local dependencies, because it combines the local modeling ability of residual convolutional layer \cite{unet} and non-local modeling ability of swin transformer \cite{SCUNet,swin_trans}, 
also cuts the computational cost due to the usage of parallel group convolution. DWT-IWT is employed to enlarge receptive field with cheap computation, meanwhile within the DWT block, DWT is used to replace each pooling operation, due to the invertibility of DWT can guarantee that such a down-sampling scheme do not introduce information loss. Both SC blocks, DWT and IWT blocks primarily contribute in reducing noise.
We train DJRD using $L_1$ loss and FFT loss: 
\begin{equation}
    L = \alpha_1 L_1(\hat{\mathbf{I}}, \mathbf{I}) + \alpha_2 \operatorname{FFT}(\hat{\mathbf{I}}, \mathbf{I}), 
\end{equation}
where $\mathbf{I}$ is the ground truth, $\alpha_1 =0.99, \alpha_2 = 0.01$.

% \begin{figure}[t!]
% \centering
% \includegraphics[width=0.45\textwidth]{images/DN_RM_first.png}
% \caption{Details of a real images. From left to right: noisy input
% (demosaicked), RM\&DN, and DN\&RM.}
% \label{flowchart}
% \end{figure}

% \begin{figure*}
% \centering
% \includegraphics[width=0.95\textwidth]{images/Slide11.jpg}
% \caption{Details of the proposed Quad Bayer CFA Convolution block.}
% \label{remosaic:attention}
% \end{figure*}

\begin{figure*}
\centering
\includegraphics[width=0.82\textwidth]{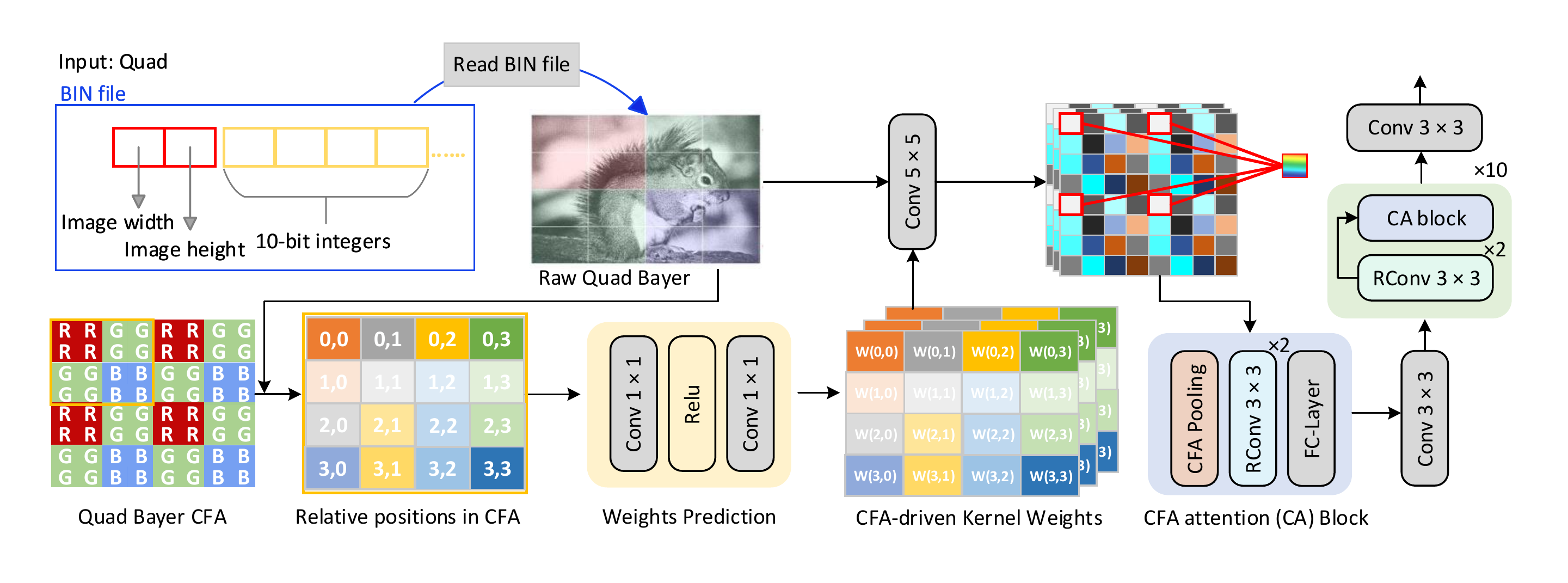}
\vspace{-5mm}
\caption{Details of the proposed CFA attention based Quad Bayer remosaicing module (QB-Re block).}
% It exploits the Quad Bayer CFA convolution and the CFA attention (CA) block as the main building block. 
% In the CFA transformer, the Quad CFA matrix is firstly reshaped into vector, and then passed into two fully-connected layers and ReLU, finally reshaped back to periodical convolution weight.
% In each CA block, the input is first passed through a CFA pooling, and subsequently is passed into a residual $3 \times 3$ convolutional
% (RConv) block and a fully-connected layers, resulting in channel feature maps, each of which is then fed into residual $3 \times 3$ RConv block and then CA block with 3 times; after that, the outputs of the refined feature maps are used reconstruct the Bayer image and then passed through a $3 \times 3$ convolution and a projection layer to produce the final remosaiced Bayer.}
\label{fig:QB-Re block}
\end{figure*}

\subsection{Quad Bayer Remosaicing Block}
 
The neighboring $2 \times 2$ cells of the raw Quad Bayer pattern always pertain to distinct channels, whereas the individual pixels within each cell are predominantly representative of a single color, as illustrated in Fig. \ref{fig:galaxy}. Consequently, applying convolution on the raw Quad Bayer data directly would lead to distorted spatial information and a loss of channel correlation. To address this issue, this subsection proposes a solution by introducing a Quad Bayer color filter array (CFA)-driven convolution block. 
% Convolutional neural networks can be directly applied to remosaic, which implicitly explores the spatial channel correlation of images. 
% However, the raw mosaiced image with single band contains the pixels from  $R, G$ and $B$ channels; adjacent $2 \times 2$ cells always belong to different channels, while the pixels inside each cell are filtered by single colour array, as shown in Fig. \ref{fig:galaxy}. 
% Nevertheless, the standard CNN only learns the spatial correlation of a single channel. 
% Therefore, applying the CNN on the raw Quad data directly will get the distorted spatial information, while losing the channel correlation. 
% Although there are some methods we can refer from demosaicing topics to alleviate it, e.g., soft rearrangement, hard splitting, or interpolation. 
% It introduces problems like reducing the resolution, resulting images with artifacts and mixing the spatial-channel information.  

Specifically, the arrangement of a Quad Bayer CFA pattern is a repeating $4 \times 4$ grid matrix containing 4 color sensitive sensor cells over the entire grid. 
To reconstruct a full-resolution Bayer image from such a Quad mosaic, we designe a CFA-aware weight sharing strategy based on CFA layout and its global periodical pattern, 
which allows the convolution kernel to change its weight periodically when sliding, so that the color distribution in the perceptual field changes periodically. 
As shown in Fig. \ref{fig:QB-Re block}, for an input Quad Bayer image $\mathbf{I} \in \mathbb{R}^{M \times N}$, its relative positions within the CFA is firstly calculated, then the $\ell$-th channel of feature map $\mathbf{F}_{1}$ is extracted by using the CFA-driven convolution kernel, in which the weights is changed periodically according to $4 \times 4$ Quad Bayer pattern,
\begin{equation}
    \begin{aligned}
    % & \mathbf{F}_{1}(i, j, k)=\operatorname{Map}\left(\mathbf{I}(i, j), 
    % \operatorname{Kernel}\left(i^{\prime}, j^{\prime}\right)\right),\\
    & \mathbf{F}_1^{\ell}=\sum_t \mathbf{K}_t^{\ell} * \mathbf{I}_t,\\
    & \mathbf{F}_1^{\ell}[i,j] = \sum_{p, q} \mathbf{I}_t[i+p, j+q] \mathbf{K}_t^{\ell}[M-1-p, N-1-q],
\end{aligned}
\end{equation}
where * denotes 2D convolution operation, $\mathbf{K}_t^{\ell}$ is kernel matrix of shape $M \times N$, which parameterizes a filter according to the relative positions of input image within Quad Bayer CFA.
It is designed to ensure that neighboring pixels with the same color have similar spectral distributions, by assigning the same weights to pixels with the same relative positions, i.e., 
\begin{equation}
    \mathbf{K}_t^{\ell}[i^{\prime}, j^{\prime}]=\operatorname{WP}\left((i^{\prime}, j^{\prime}) ; \theta\right),
\end{equation}
where $\left(i^{\prime}, j^{\prime}\right)=(i \bmod 4, j \bmod 4)$ is the relative position of this pixel in the Quad Bayer CFA, $\operatorname{WP}$ is the weight prediction block that predicts the kernel weights by using $\left(i^{\prime}, j^{\prime}\right)$.
Then, feature $\mathbf{F}_1$ is processed by a CFA attention (CA) block, which consists of a CFA pooling layer, two $3 \times 3$ residual convolution (RConv) and a fully-connected layer. 
Specifically, for $\mathbf{F}_1 \in \mathbb{R}^{M \times N \times C}$, CFA pooling aggregates the feature points with the same relative position,
\begin{equation}
    \mathbf{F}_2\left(i, j, k)\right)=c \sum_{s=0}^{\frac{M}{4}-1} \sum_{t=0}^{\frac{N}{4}-1} \mathbf{F}_1[i+4s, j+4t, k],
\end{equation}
where $c=\frac{1}{M / 4 \times N / 4}$. Then, two $3 \times 3$ RConv are used to extract attention map $\mathbf{F}_A$,
\begin{equation}
    \mathbf{F}_A = \operatorname{RConv}(\operatorname{RConv}(\mathbf{F}_2)),
\end{equation}
and then $\mathbf{F}_A$ is refined by the fully-connected layer.
Subsequently, the attention map $\mathbf{F}_A$ further passes through a $\operatorname{Conv}3\times3$ layer. Finally, we stack CA blocks together with two residual convolution layers 10 times, and the output of QB-Re block is generated by using a $\operatorname{Conv}3\times3$ layer. 

Distinguished from CNNs that share the global weights, the proposed QB-Re block allocates different weights to channels with varied colors, by using a CFA-aware weight. 
To further reduce spatial information loss, a CFA attention module (CA) is also proposed, in which we employ a CFA-sensitive mechanism to aggregate the features, called CFA pooling. 
It aggregates the features within the same relative position in the CFA, which enables the CA to focus on loading of CFA patterns within each channel.

\subsection{Dual-head Joint Remosaic and Denoise}

In this subsection, we improve the practicality of remosaicing model for real images that are degraded by mixed noise, by integrating a denoising block into the network through a plug-and-play structure, i.e., remosaicing-denoising. This integration makes the network flexible as denoising modules can be independently improved through pre-training or network design.

Subsequently, we evaluated the advantages and disadvantages of solving the denoising and remosaicing (DN\&RM) problem in both the DN\&RM and RM\&DN orders. In the RM\&DN order, the noise loses its independent identically distributed (i.i.d.) property, becoming more complex after the raw image is processed by the remosaic module. Consequently, denoising modules that rely on the i.i.d. assumptions become less effective. On the other hand, the DN\&RM order does not involve handling a raw image with complicated noise, making denoising easier to implement. However, some details may be lost in the denoised image due to the absence of a perfect denoising algorithm, which may be amplified by subsequent remosaicing.

To address this issue, we propose a parallel solution with dual-head by integrating both DN\&RM and RM\&DN strategies together, thus avoiding the issue of determining which step should be performed first. However, the inputs passed into the denoising modules in the two schemes are quite different: one of them is the remosaiced Bayer image with noise, and the other one is the noisy Quad Bayer image. To address this, we customize two specific denoising blocks for each scheme, as depicted in Fig. \ref{remosaic:sketch_U}. 

Specifically, for the first branch, the input is the noisy Quad Bayer mosaic. As adjacent $2 \times 2$ pixels in the Quad sensor often come from the same channel in $R, G$, and $B$, the pixels in adjacent $2 \times 2$ squares often have different colors. Therefore, using CNNs that represent as much local information as possible may not be sufficient for effective denoising. Here, we consider both local and non-local information by employing a swin-transformer integrated residual convolution (SC) block \cite{SCUNet}, and stack it in a multiscale UNet style. This approach incorporates the local modeling capability of the residual convolution and the non-local modeling capability of the swin transformer, resulting in effective denoising of the Quad Bayer mosaic,
\begin{eqnarray}
  \left\{
    \begin{aligned}  
    & \mathbf{Y}_0 = \operatorname{Conv}3\times3(\mathbf{O}) \in \mathbb{R}^{M \times N \times C} \\
    & \mathbf{Y}_l = \operatorname{DownS} (\operatorname{SC}(\mathbf{Y}_0)) \in  \mathbb{R}^{\frac{M}{8} \times \frac{N}{8} \times 8C } \\ 
    & \mathbf{Y}_{l+1}  = \operatorname{SC}(\mathbf{Y}_{l}) \in \mathbb{R}^{\frac{M}{8} \times \frac{N}{8} \times 8C } \\
    & \mathbf{Y}_{l+2}  = \operatorname{UpS} (\operatorname{SC}(\mathbf{Y}_{l+1})) \in \mathbb{R}^{M \times N \times C} \\
    &\mathbf{Y}_{l+3} = \operatorname{Conv}3\times3(\mathbf{Y}_{l+2}) \in \mathbb{R}^{M \times N} \\
    &\mathbf{Y}_{l+4} = \operatorname{QB-Re} (\mathbf{Y}_{l+3}, \mathbf{Q}) \in \mathbb{R}^{M \times N}
    \end{aligned}
  \right.
\end{eqnarray}
where $\operatorname{DownS, UpS}$ are the down-sampling ($2\times2$ strided convolution with stride 2) and up-sampling ($2\times2$ transposed convolution with stride 2), respectively.

For the second branch, the input of DN is the Bayer image processed by QB-Re block, it has small CFA pattern, the $2 \times 2 $ cell includes pixels from $R, G$ and $B$ channels, and adjacent pixels often belong to different channels, instead of one color for one cell in the Quad image.
To effectively increase the perceptual field of view of pixels belonging to different colors, avoiding be entrapped into local CFA patterns, while taking into account the computational power of mobile devices such as mobile phones, we employ DWT-IWT blocks,
\begin{eqnarray}
  \left\{
    \begin{aligned}
    & \mathbf{X}_l = \operatorname{QB-Re} (\mathbf{O}, \mathbf{Q}) \in \mathbb{R}^{M \times N} \\
    & \mathbf{X}_{l+1} = \operatorname{DWT}(\mathbf{X}_l) \in \mathbb{R}^{\frac{M}{8} \times \frac{N}{8} \times 8C } \\
    & \mathbf{X}_{l+2} = \operatorname{PReLu}(\operatorname{Conv}(\mathbf{X}_{l+1}) )\in \mathbb{R}^{\frac{M}{8} \times \frac{N}{8} \times 8C } \\
    & \mathbf{X}_{l+2} = \operatorname{RG} (\operatorname{RG}(\mathbf{X}_{l+1}) )\in \mathbb{R}^{\frac{M}{8} \times \frac{N}{8} \times 8C } \\
    & \mathbf{X}_{l+3} = \operatorname{IWT}(\mathbf{X}_{l+2}) \in \mathbb{R}^{M \times N}\\
    & \mathbf{X}_{l+4} = \operatorname{PReLu}(\operatorname{Conv}(\mathbf{X}_{l+3}) )\in \mathbb{R}^{M \times N} \\
    \end{aligned}
  \right.
\end{eqnarray}
Moreover, DWT can capture both frequency and location information of feature maps \cite{daubechies1992ten,daubechies1990wavelet}, which is also helpful to reduce information loss during denoising phase and feed the subsequent QB-Re block with more information. Subsequently, $\hat{\mathbf{I}}$ is obtained by aggregating $\mathbf{X}_{l+4}$ and $\mathbf{Y}_{l+4}$.

\begin{figure}[!htbp]
	% \captionsetup{font=small}
	\centering
	% \scriptsize
	\renewcommand{\h}{0.105}
	\renewcommand{\wa}{0.12}
	\newcommand{\wb}{0.16}
	\renewcommand{\g}{-0.7mm}
	\renewcommand{\tabcolsep}{1.8pt}
	\renewcommand{\arraystretch}{1}
        \resizebox{1.0\linewidth}{!} {
		\begin{tabular}{cc}			
			\renewcommand{\name}{images/hard_case_demo/}
			\renewcommand{\h}{0.15}
			\renewcommand{\w}{0.2}
			\begin{tabular}{cc}
				% \begin{adjustbox}{valign=t}
				% 	\begin{tabular}{c}%
		  %        	\includegraphics[trim={0 0 180 0 },clip, width=0.275\textwidth]{\name gt/img_007_SRF_2_HR_0db.png}
				% 		\\
				% 		Urban100-007 
				% 	\end{tabular}
				% \end{adjustbox}
				\begin{adjustbox}{valign=t}
					\begin{tabular}{cccccc}
						\includegraphics[trim={1300 600 100 300  },clip,height=\h \textwidth, width=\w \textwidth]{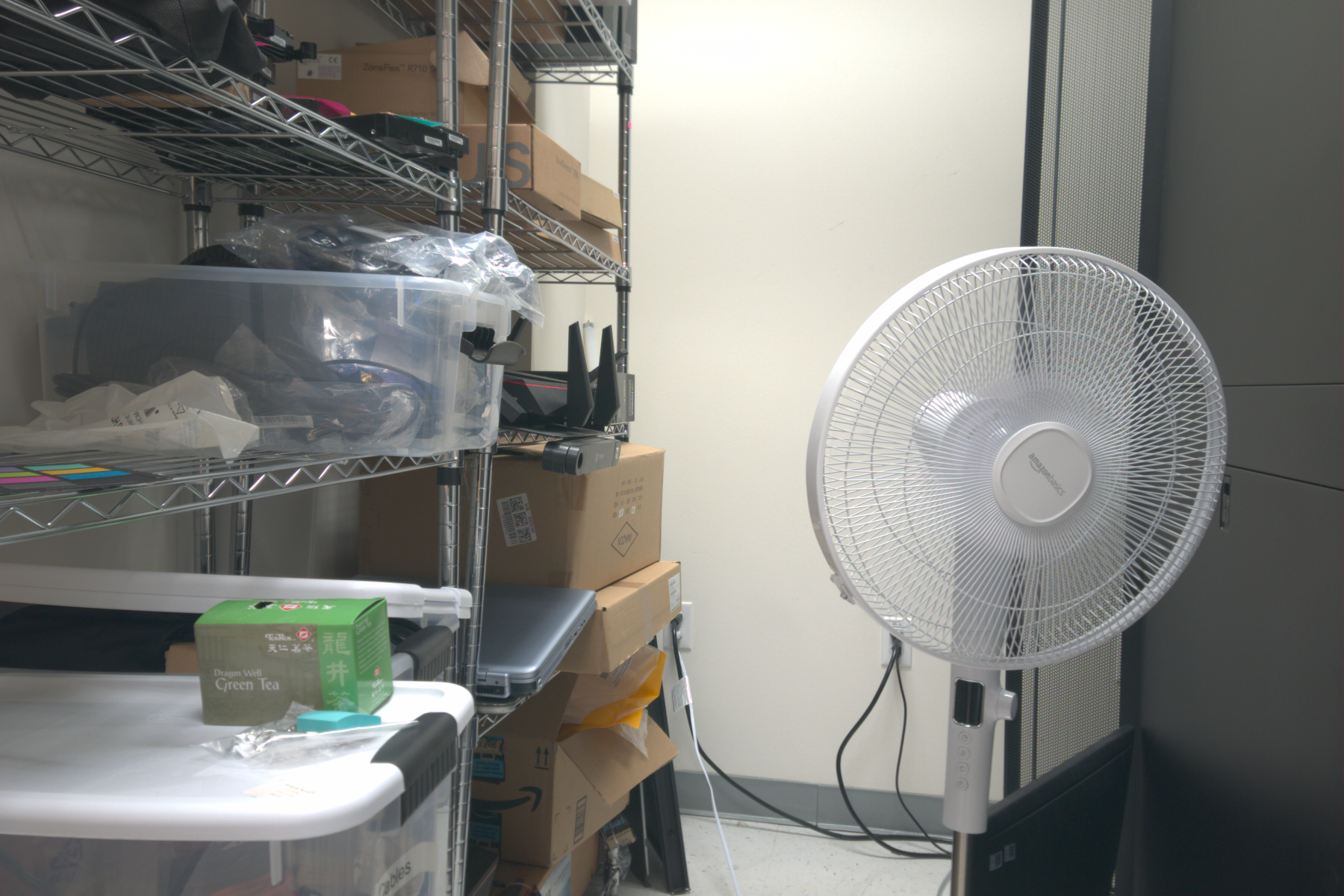} \hspace{\g} &
						\includegraphics[trim={1300 600 100 300  },clip,height=\h \textwidth, width=\w \textwidth]{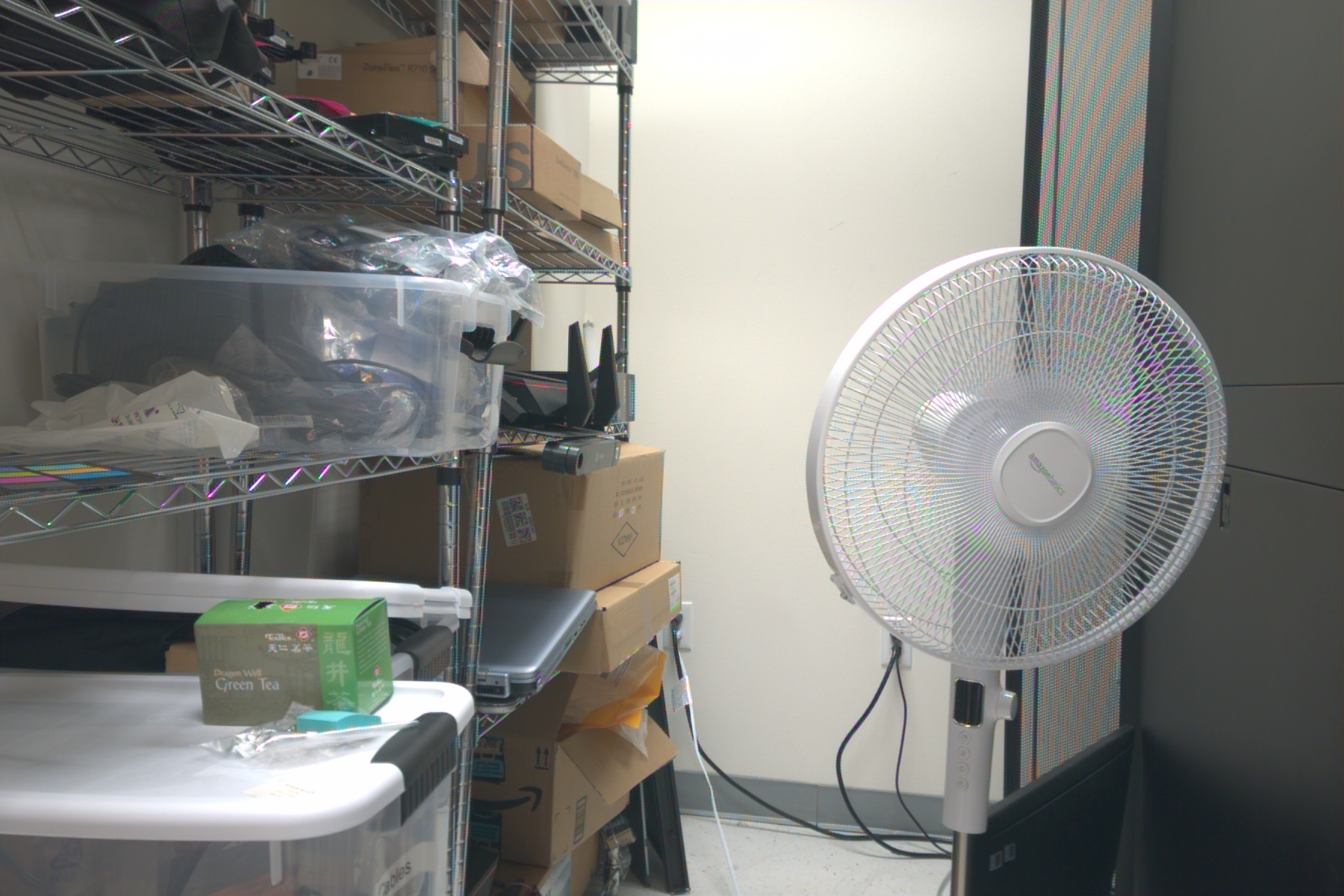} \hspace{\g} &
						\includegraphics[trim={1300 600 100 300   },clip,height=\h \textwidth, width=\w \textwidth]{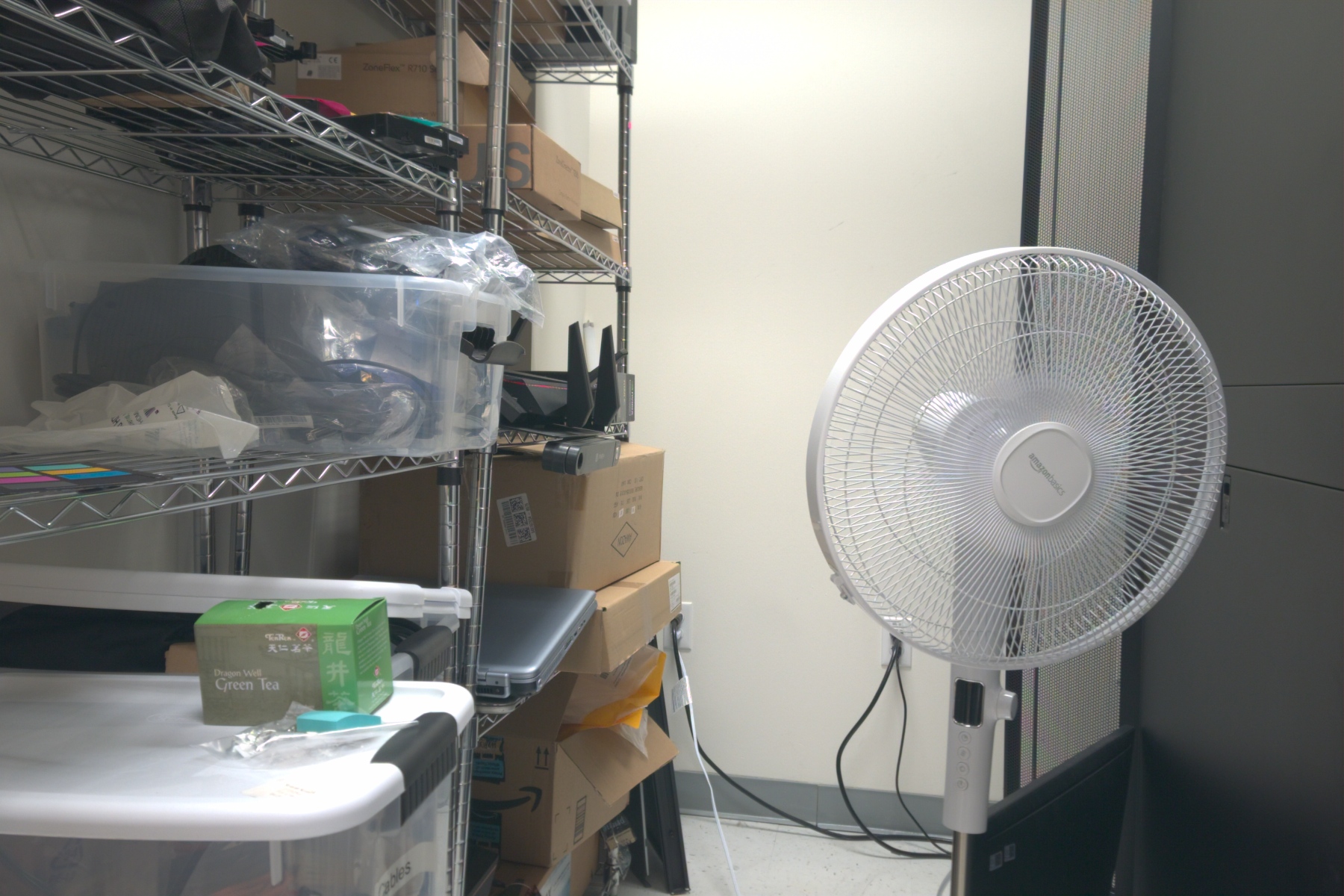}
						\\
						GT \hspace{\g} & \textbf{W/O} fine-tuning  \hspace{\g} &	\textbf{W/} fine-tuning
						\\
						\vspace{-3.5mm}
						\\
						\includegraphics[trim={820 1100 900 60   },clip,height=\h \textwidth, width=\w \textwidth]{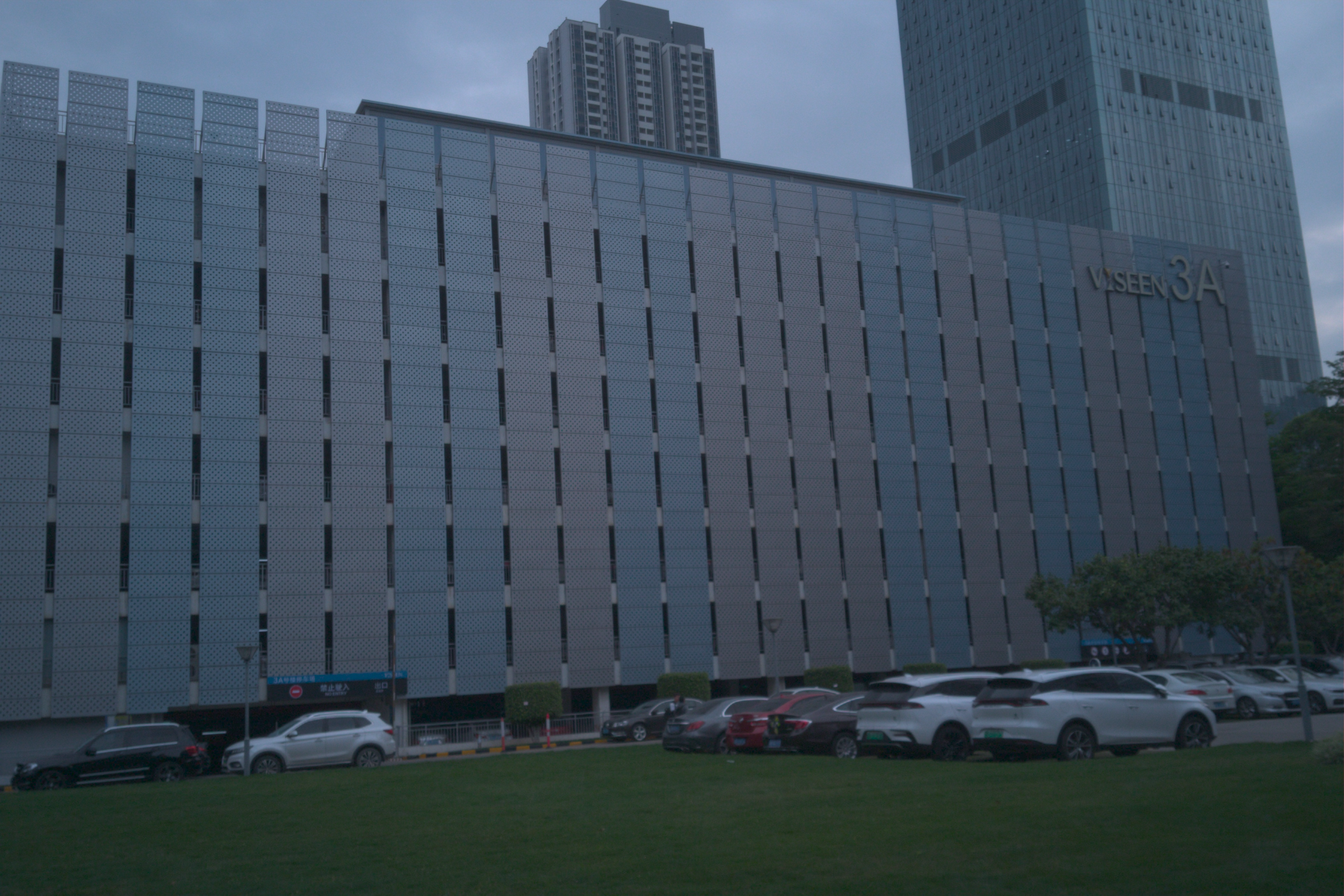} \hspace{\g} &
						\includegraphics[trim={820 1100 900 60   },clip,height=\h \textwidth, width=\w \textwidth]{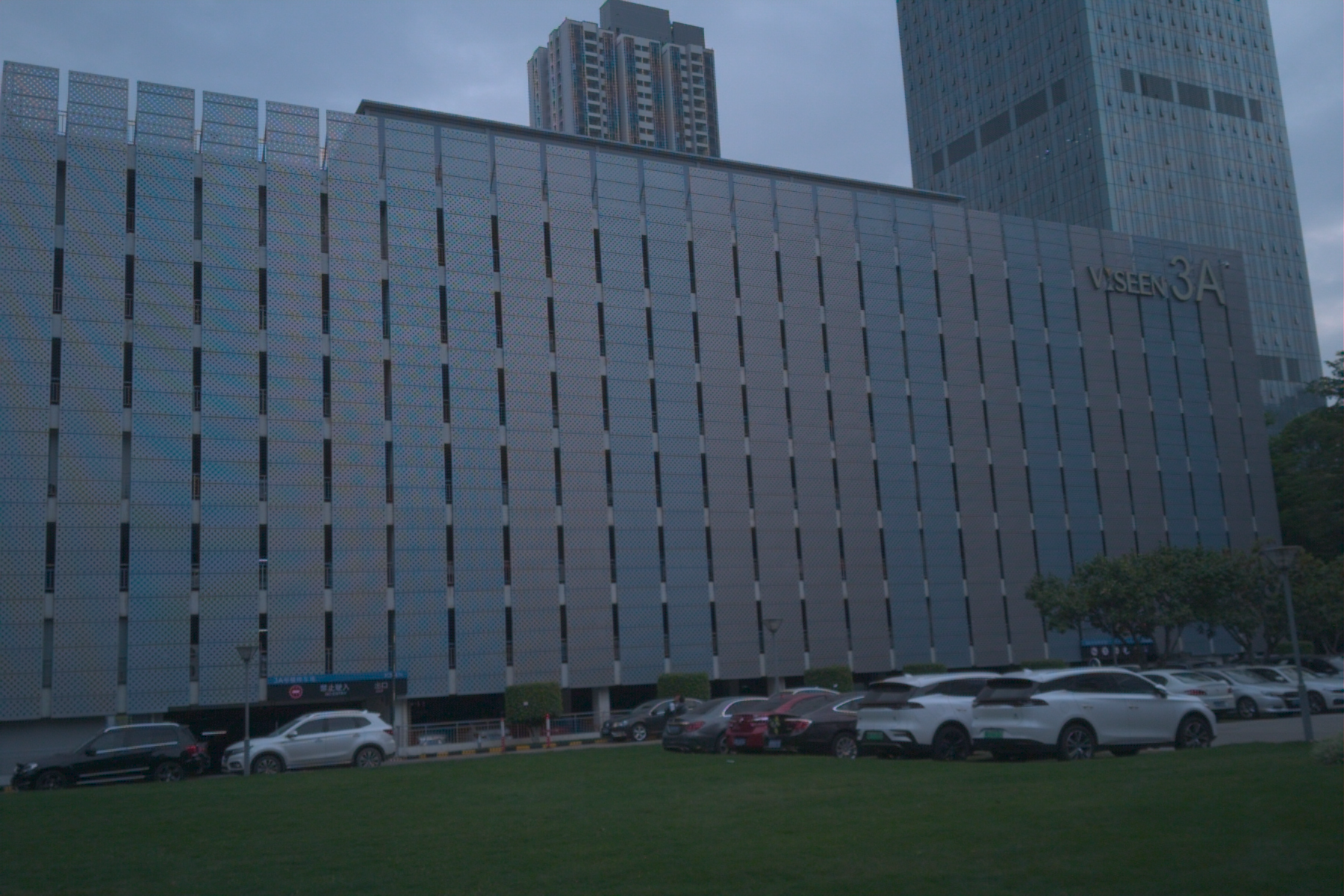} \hspace{\g} &
						\includegraphics[trim={800 1100 900 60   },clip,height=\h \textwidth, width=\w \textwidth]{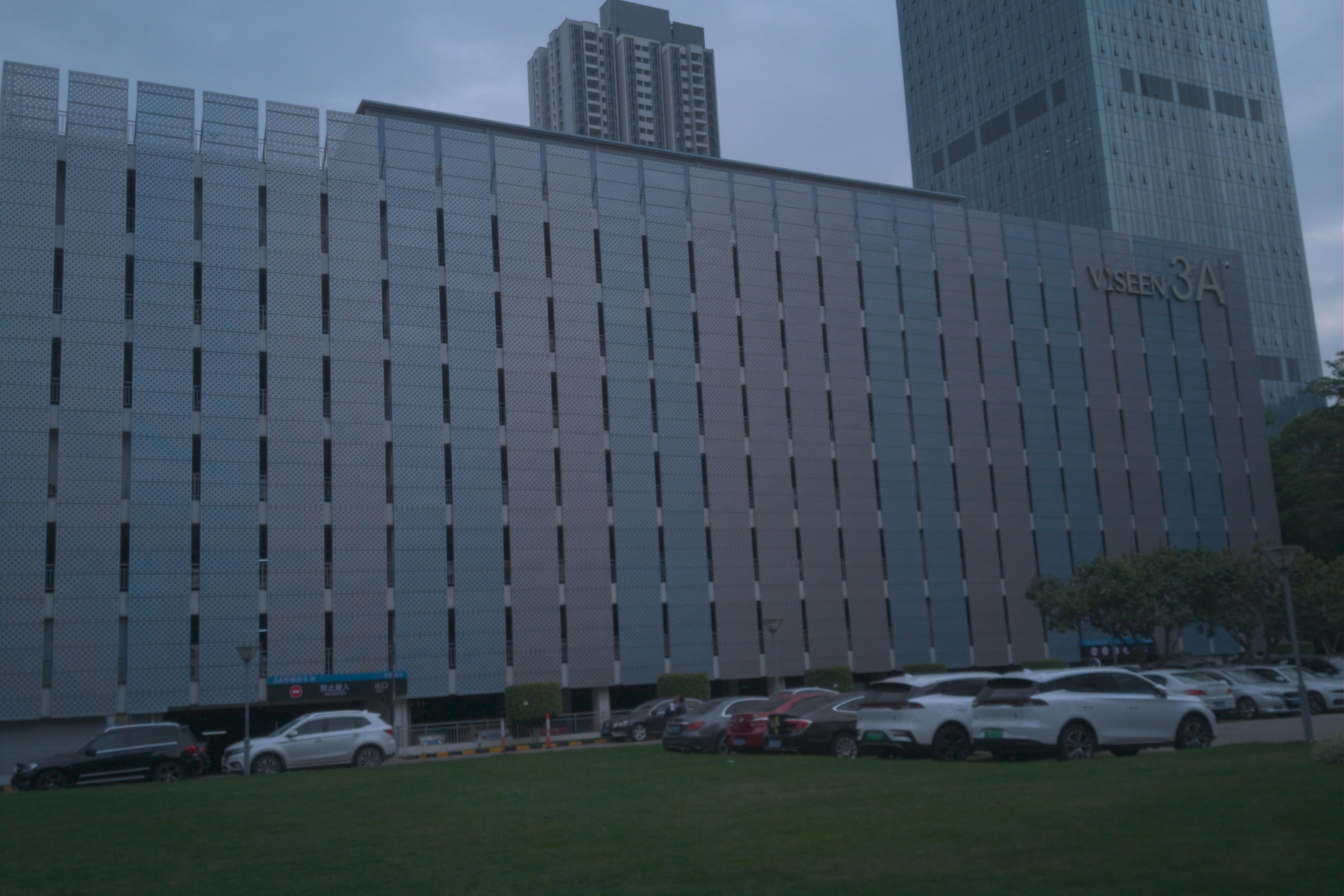} % 500 900 1200 225
						\\
						GT \hspace{\g} & \textbf{W/O} fine-tuning  \hspace{\g} &	\textbf{W/} fine-tuning
						\\
					\end{tabular}
				\end{adjustbox}
			\end{tabular}	
		\end{tabular}
	}
	\vspace{-2mm}
	\caption{After training the proposed DJRD on a standard image dataset, we observed noticeable artifacts such as zippering on thin building windows and Moiré patterns in the fan example. However, fine-tuning the network on challenging cases significantly reduced these artifacts, as seen in the second column.} %
	  \vspace{-2mm}
	\label{fig:HardCases}
\end{figure}

% \begin{figure*}[t!]
% \centering
% \includegraphics[width=0.95\textwidth]{images/sketch_whole_model.png}
% \caption{Overview of the training procedure for fine-tuning on hard cases.}
% \label{fig:sketch}
% \end{figure*}

\subsection{Bottleneck Data Mining} \label{HardCase}

% \noindent \textbf{Training Data Synthesis}: the publicly available remosaic dataset contains hundreds of images, not enough to train the deep network with thousands of parameters. 
% Therefore, we first download additional RGB images from imagenet and obtain pre-calibration RGB images by reversing the camera's ISP, as shown in the top of Fig. \ref{fig:sketch}. 
% Then, we generate a mosaic and noisy image by keeping only one color channel per pixel, based on Bayer pattern and Quad Bayer CFA. 
% The Bayer and Quad image pairs generated by this strategy fit well with the camera ISP settings.

% \noindent \textbf{Addressing bottleneck of SOTA}: 
On the most general test datasets, most of existing CNN methods can recover images that are visually close to the ground truth. 
However, when we zoom in locally, a closer inspection reveals artifacts near fine edges and complex textures (see Fig.~\ref{fig:HardCases}). 
This suggests that a large number of training samples does not guarantee convincing re-mosaicing. 
This is mainly caused by the distributional properties of training data. 
Specifically, the randomly selected images are mainly composed of smooth blocks, as these blocks dominate the natural images \cite{levin2012patch,MIT_deep_joint}. 
Therefore, with such a training dataset challenging structures account for only a small fraction, smoothed patches occupy the vast majority of the training data, when its number reaches a certain value, the performance improvement brought by continuing to add such training samples is tiny. 
% In general, it is necessary to compensate training images in the tails of the data distribution in a targeted manner to further improve the performance of the model on difficult examples and thus break the existing performance bottleneck.

To overcome the bottleneck problem, we rebuild a \emph{Bottleneck dataset}. This dataset comprises 2000 $128\times128$ hard patches, which include paired Quad Bayer, Bayer, and RGB images. The dataset creation process begins by utilizing our DJRD model, trained on the MIPI dataset, to acquire Bayer images. We then convert these images to the RGB domain using the demosaicing method MIT \cite{MIT_deep_joint}. Subsequently, we select a database that contains images degraded by two specific artifacts, namely zipper and color Moiré, as illustrated in the second column of Fig.~\ref{fig:HardCases}. Specifically, we employ the HDR-VDP2 visual metric \cite{MIT_deep_joint,HDR_VDP} to identify hard cases featuring zipper artifacts along thin edges.
% It has been shown that HDR-VDP2 can effectively model human visual systems, by computing the response of local artifacts and global image quality \cite{MIT_deep_joint}.
Then, for the Moiré (as the distracting false color bands shown in the fan of Fig.~\ref{fig:HardCases}), it is caused by the misaliasing of adjacent pixels belonging to different color channels and introduces undesirable low frequency textures.
% erroneous interpolation of color samples. 
% The adjacent pixels in the raw mosaic images always belong to different colors, which results in mis-aliasing and introduces undesirable low frequency textures.
Therefore, we measure the frequencies by quantifying the difference of each frequency as follows:
\begin{equation}
    \rho(\omega)= \begin{cases}\log \left(\frac{\left|\mathcal{F}_{\mathrm{CI}}(\omega)\right|^2+\eta}{\left|\mathcal{F}_{\mathrm{GT}}(\omega)\right|^2+\eta}\right) & \text { if }|\omega| \leq c \\ 1 & \text { otherwise }\end{cases}
\end{equation}
where $\mathcal{F}_{\mathrm{CI}}(\omega)$ and $\mathcal{F}_{\mathrm{GT}}(\omega)$ denote the 2D Fourier transform of each channel of the compared image (CI) and ground truth (GT), respectively. $c = 0.95 \pi$ is a constant. 
% used to mitigate boundary effects and high-frequency noise.
% see \cite{MIT_deep_joint} for more details. 

With the Bottleneck dataset, the loss function is then effectively reweighed toward difficult patches, by focusing on fitting the hard images while rejecting trivial cases. 
Fig. \ref{fig:HardCases} illustrates an example of training our network on hard cases, which shows that the reweighed network yielded drastically improved results, especially the zipper and Moir\'e artifacts. 
% More results are to be seen in Section \ref{ablation_section}.

\section{Experiments}

% In this section, the superiority of the proposed mode is evaluated on three public datasets in both Bayer domain and sRGB domain. 
% Ablation experiments are also implemented to demonstrate the practicability of the key modules of the proposed algorithm.

\subsection{Datasets and Implementation Detail}

To better test the performance of the proposed model, we use 210 images with size of $1200 \times 1800$, from the latest 2022 MIPI challenge \cite{yang2022mipi}, as the basic training set. 
The training images for all the tested models have three noise levels: 0dB, 24dB and 42dB, all the noise consists of read noise and shot noise.
% The detailed training procedure is provided in the top of Fig. \ref{fig:sketch}.
Additionally, the selected hard cases is used to fine-tune the trained model.
For equal comparison, in testing, we also use the standard test datasets released by the challenge, which contains 30 images with size of $1200 \times 1800$.
In addition, two public image datasets: Urban100 \cite{urban100} and MIT Moiré \cite{MIT_deep_joint} are chosen as test images. 
MIT Moiré images consist of 210 images, and the whole Urban100 has 100 high-resolution images. 
% The training and test experiments are implemented in PyTorch.
% $l_1$, $l_2$ and FFT loss are used for all the modules.

\subsection{Results in Bayer and sRGB Domain}

Firstly, it should be noted that there is currently no publicly available full-resolution remosaicing model. Therefore, we evaluate the proposed remosaic model DJRD on the Bayer domain separately. We use the probability distribution difference based Kullback-Leibler divergence (KLD), Peak Signal-to-Noise Ratio (PSNR), and Learned Perceptual Image Patch Similarity (LPIPS) \cite{LIPIS}. Tab.~\ref{Table:PQI_KLD} shows that our model produces high-quality Bayer images, with a PSNR of over 40 dB and a KLD smaller than 0.025.
% To make the comparison with SOTA Quad demosaicing models in the RGB domain, a simple ISP is used to convert the remosaiced Bayer to RGB.

% For training, the remosaic and denoise modules are trained separately, and then we retrain them together. 
% For all the models we test here, the basic training images are used to train them firstly. 
% Then the created hard cases are used to fine tune the deep models.
% We notice that retraining the BJDD and SAGAN with hard datasets, does not improve its performance, even infects the accuracy. 
% The reason may be that these two models rely on joint network architecture, which integrates the denoise and demosaic module in one network. 
% In this way, such a network prefers training images with noise, the noise free difficult patches may contribute to the demosaic part, but does not feed any valuable information to the denoise module.

% In addition, for the denoising module, we employ the SCUNet \cite{SCUNet} and DWT \cite{2020ntire_denoising} to address the realistic noise in this task. 
% The SCUNet is used to denoise the Quad Bayer image, and we do not use the pre-trained mode, and only train it on the provided images. 
% The DWT is used to denoise the remosaiced Bayer image.

\begin{table}[!htbp]
\centering
\renewcommand{\arraystretch}{1.2}%
\caption{Quantitative evaluation in Bayer domain, sRGB domain.}
\label{Table:PQI_KLD}
\scalebox{0.73}{
\begin{tabular}{cc|ccc|ccc}
% \Xcline{1-8}{0.9pt}
\multirow{2}{*}{Dataset} &\multirow{2}{*}{Metric}&	\multicolumn{3}{c}{Bayer Domain} & \multicolumn{3}{c}{sRGB}\\
\Xcline{3-8}{0.4pt}
& &	0dB 	&	24dB 	&	42dB & 0dB&24dB& 42dB\\
\Xcline{1-8}{0.4pt}
\multirow{3}{*}{MIPI}  & KLD &  0.0037 & 0.0096& 0.0237 & -& -&-\\
% & Urban100  & 0.2105 & -& -\\
% & MIT MOIRE  & 0.4107 & -& -\\
& PSNR  & 51.51 & 45.31& 40.45 &40.58 &36.19 &32.43\\
% & SSIM  & 0.4107 & -& -\\
& LPIPS  & 0.0034 & 0.0579 & 0.1366 &0.0301 &0.1316 &0.2305\\
% \Xcline{1-8}{0.9pt}
\end{tabular}}
\end{table}

Subsequently, the proposed model was compared to state-of-the-art Quad Bayer demosaicing methods in the sRGB domain, including the classical joint demosaicing and denoising model \cite{MIT_deep_joint} (referred to as MIT), the latest deep attention-based PIPNet \cite{BJDD}, and SAGAN, which employs adversarial spatial-asymmetric attention \cite{SAGAN}. To enable the comparison in the sRGB domain, we used the pre-trained MIT \cite{MIT_deep_joint} to convert our Bayer images generated by DJRD to sRGB images.
\textbf{Quantitative Results}: The proposed model was evaluated using three image quality metrics: PSNR, SSIM, and LPIPS. Tab.~\ref{Table:PQI_urban_MIT}, \ref{Table:PQI} present the reconstructed results of all the test modes on images with three noise levels. The results show that our DJRD produces the best overall performance. Specifically, DJRD outperforms PIPNet and SAGAN with 5.68dB and 4.54dB PSNR, respectively. Additionally, there is a 0.04 SSIM gap between DJRD and SAGAN.
% Interestingly, while the proposed model has an obvious improvement over PIPNet and SAGAN under the three noise levels, it achieves a larger improvement with lower noise level. e.g., the gap between our model and SAGAN, with 0dB, 24dB , 42dB inputs are 7.05dB, 3.77dB and 2.8dB, respectively. 
% This is maybe because empowered by the proposed joint remosaic and denoise network, our model converts the raw Quad to Bayer with high quality. 
% In addition, under high level practical mixed noise the raw Quad image is degraded severely, consequently, there is an obvious quality drop among all the tested models.

% \begin{figure}[ht!]
%     \centering
%     \includegraphics[width=0.47\textwidth]{images/ECCV_comparison_4.jpg}
%     \caption{An image of comparison.}
%     \label{fig: Compare_4}
% \end{figure}

% \begin{figure*}[ht!]
%     \centering
%     \includegraphics[width=0.85\textwidth]{images/ECCV_comparison_2.jpg}
%     \caption{An image of comparison.}
%     \label{fig: Compare_2}
% \end{figure*}

% \begin{figure*}[ht!]
%     \centering
%     \includegraphics[width=0.47\textwidth]{images/ECCV_comparison_3.jpg}
%     \caption{An image of comparison.}
%     \label{fig: Compare_3}
% \end{figure*}

\begin{table}[!htbp]
\centering
% \tabcolsep{1.0}
\renewcommand{\arraystretch}{1.2}%
\caption{Quantitative comparison with respect to PSNR, SSIM and LPIPS in sRGB domain on MIT Moiré and Urban100.}
\label{Table:PQI_urban_MIT}
\scalebox{0.78}{
\begin{tabular}{c|cccc}
% \Xcline{1-5}{0.9pt}
\specialrule{0em}{1pt}{1pt}
Datasets &Method&	PSNR $\uparrow$	&	SSIM $\uparrow$	&	LPIPS $\downarrow$ \\
% \specialrule{0em}{1pt}{1pt}
\Xcline{1-5}{0.4pt}
\multirow{4}{*}{Urban100 \cite{urban100}} &MIT \cite{MIT_deep_joint}	&	24.87	&	0.90	&	0.1217	\\
&PIPNet \cite{BJDD}	&	26.67	&	0.93	&	0.0951	\\
&SAGAN \cite{SAGAN}	&	26.34	&	0.92	&	0.1067	\\
& \textbf{DJRD}(Our)	&	\textbf{31.02}	&	\textbf{0.98}	&	\textbf{0.0348}	\\
\Xcline{1-5}{0.4pt}
\multirow{4}{*}{MIT Moiré \cite{MIT_deep_joint}} &MIT \cite{MIT_deep_joint}	&	25.20	&	0.81	&	0.1857	\\
&PIPNet \cite{BJDD}	&	26.04	&	0.88	&	0.1740	\\
&SAGAN \cite{SAGAN}	&	26.42	&	0.87	&	0.2034	\\
& \textbf{DJRD}(Our)	&	\textbf{29.74}	&	\textbf{0.95}	&	\textbf{0.0815}	\\
% \Xcline{1-5}{0.9pt}
\end{tabular}}
\end{table}

\begin{figure}[!htbp]
    % \centering
    \scriptsize
    \renewcommand{\name}{images/mire/}
    \includegraphics[width=0.09\textwidth]{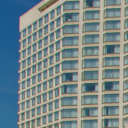}
    \hspace{-3.5mm}
    \includegraphics[width=0.09\textwidth]{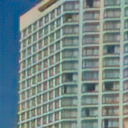}
    \hspace{-3.5mm}
    \includegraphics[width=0.09\textwidth]{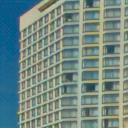}
    \hspace{-3.5mm}
    \includegraphics[width=0.09\textwidth]{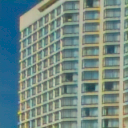}
    \hspace{-3.5mm}
    \includegraphics[width=0.09\textwidth]{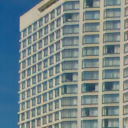}
    \vspace{-3.5mm}
    \includegraphics[width=0.09\textwidth]{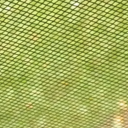}
    \hspace{-3.5mm}
    \includegraphics[width=0.09\textwidth]{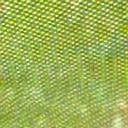}
    \hspace{-3.5mm}
    \includegraphics[width=0.09\textwidth]{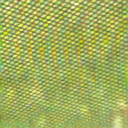}
    \hspace{-3.5mm}
    \includegraphics[width=0.09\textwidth]{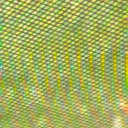}
    \hspace{-3.5mm}
    \includegraphics[width=0.09\textwidth]{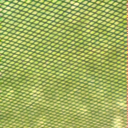}
    \vspace{-1mm}
    \subfigure[GT]{\includegraphics[width=0.09\textwidth]{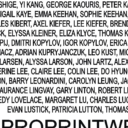}}
    \hspace{-0.06mm}
    \subfigure[MIT]{\includegraphics[width=0.09\textwidth]{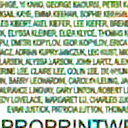}}
    \hspace{-0.06mm}
    \subfigure[PIPNet]{\includegraphics[width=0.09\textwidth]{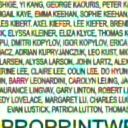}}
    \hspace{-0.06mm}
    \subfigure[SAGAN]{\includegraphics[width=0.09\textwidth]{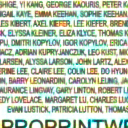}}
    \hspace{-0.06mm}
    \subfigure[\textbf{DJRD}]{\includegraphics[width=0.09\textwidth]{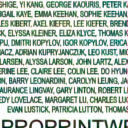}}
    \vspace{-0.2mm}
    \caption{Visual comparison on MIT Moiré and Urban100.}
    \label{fig: moire}
\end{figure}

\noindent\textbf{Visual Evaluation}:
% \begin{figure*}[ht!]
%     \centering
%     \includegraphics[width=0.85\textwidth]{images/ECCV_comparison_1.jpg}
%     \caption{An image of comparison.}
%     \label{fig: Compare_1}
% \end{figure*}
% On the basis of quantitative evaluation, 
The visual results are shown in Fig.~\ref{fig: moire}, \ref{fig_mipi}, \ref{fig_urban_07}. 
For all the test models, the PSNR of its reconstructed images are over or near 30dB, which means that it is hard to observe obvious differences in a coarse scale. 
Therefore, to highlight the difference, we enlarge the challenging patches with rich details. 
% \begin{figure}
%     \centering
%     \includegraphics[width=0.48\textwidth]{images/urban_new.jpg}
%     \caption{Qualitative evaluation on Urban100 dataset.}
%     \label{fig: urban}
% \end{figure}
% \begin{figure}
%     \centering
%     \includegraphics[width=0.48\textwidth]{images/Moire - Urban.jpg}
%     \caption{Qualitative evaluation on Urban100 and MIT Moiré dataset.}
%     \label{fig: moire}
% \end{figure}
From the figures, one can see that most models suffer from residual noise or blurring artifacts. 
Specifically, Fig.~\ref{fig_mipi} provides the visual results on MIPI dataset with noise level 42dB. 
In this case the gap between the proposed mode and other methods is obvious, for example, one can observe that both PIPNet and SAGAN fail to recover the net structure, also introducing some distortion in the joints.
In contrast, the proposed mode recovers fine structures and preserves the texture in the joints.
Fig.~\ref{fig: moire}, \ref{fig_urban_07} show the results on Urban100 and MIT Moiré datasets, one can see that the wall, net and texts reconstructed by our mode preserves more details than PIPNet and SAGAN, which introduce some smoothness. More detailed analysis and discussion please refer to the \textbf{supplementary material}.

\begin{table*}[!htbp]
\centering
\setlength{\tabcolsep}{4pt}
\renewcommand{\arraystretch}{1.2}%
\caption{Quantitative comparison with respect to PSNR, SSIM and LPIPS on test image dataset MIPI.}
\label{Table:PQI}
\scalebox{0.70}{
\begin{tabular}{cc|ccc|ccc|ccc|ccc}
% \Xcline{1-14}{0.9pt}
 \multicolumn{2}{c|}{Noise Level}  &	\multicolumn{3}{c|}{0dB}	&	 		\multicolumn{3}{c|}{24dB}		&	\multicolumn{3}{c|}{42dB} 		&	\multicolumn{3}{c}{Average}	\\
Dataset &Method&	PSNR $\uparrow$	&	SSIM $\uparrow$	&	LPIPS $\downarrow$	&	PSNR $\uparrow$	&	SSIM $\uparrow$	&	LPIPS $\downarrow$	&	PSNR $\uparrow$	&	SSIM $\uparrow$	&	LPIPS $\downarrow$ &	PSNR $\uparrow$	&	SSIM $\uparrow$	&	LPIPS $\downarrow$\\
\Xcline{1-14}{0.4pt}
\multirow{4}{*}{MIPI \cite{yang2022mipi}}&MIT \cite{MIT_deep_joint}	&	30.39	&	0.89	&	0.1437	&	29.92	&	0.87	&	0.2024	&	27.93	&	0.79	&	0.3501 & 29.41 & 0.85 & 0.2321\\
&PIPNet \cite{BJDD}	&	32.32	&	0.95	&	0.1252	&	31.26	&	0.92	&	0.1800	&	28.44	&	0.87	&	0.2928 & 30.67 & 0.91 & 0.1993\\
&SAGAN \cite{SAGAN}	&	33.36	&	0.95	&	0.1161	&	32.40	&	0.92	&	0.1705	&	29.66	&	0.87	&	0.2714 & 31.81& 0.91& 0.1860 \\
& \textbf{DJRD(Ours)}	&	\textbf{40.58}	&	\textbf{0.97}	&	\textbf{0.0301}	&	\textbf{36.19}	&	\textbf{0.93}	&	\textbf{0.1316}	&	\textbf{32.43}	&	\textbf{0.89}	&	\textbf{0.2305}  & \textbf{36.40}& \textbf{0.93} & \textbf{0.1307}\\
% \Xcline{1-14}{0.9pt}
\end{tabular}}
\end{table*}

\begin{figure*}[!htbp]
	% \captionsetup{font=small}
	\centering
	\scriptsize
	\renewcommand{\h}{0.105}
	\renewcommand{\wa}{0.12}
	\newcommand{\wb}{0.16}
	\renewcommand{\g}{-0.7mm}
	\renewcommand{\tabcolsep}{1.8pt}
	\renewcommand{\arraystretch}{1}
        \resizebox{0.88\linewidth}{!} {
		\begin{tabular}{cc}			
			\renewcommand{\name}{images/}
			\renewcommand{\h}{0.13}
			\renewcommand{\w}{0.2}
			\begin{tabular}{cc}
				\begin{adjustbox}{valign=t}
					\begin{tabular}{c}%
		         	\includegraphics[trim={625 10 100 70 },clip, width=0.285\textwidth]{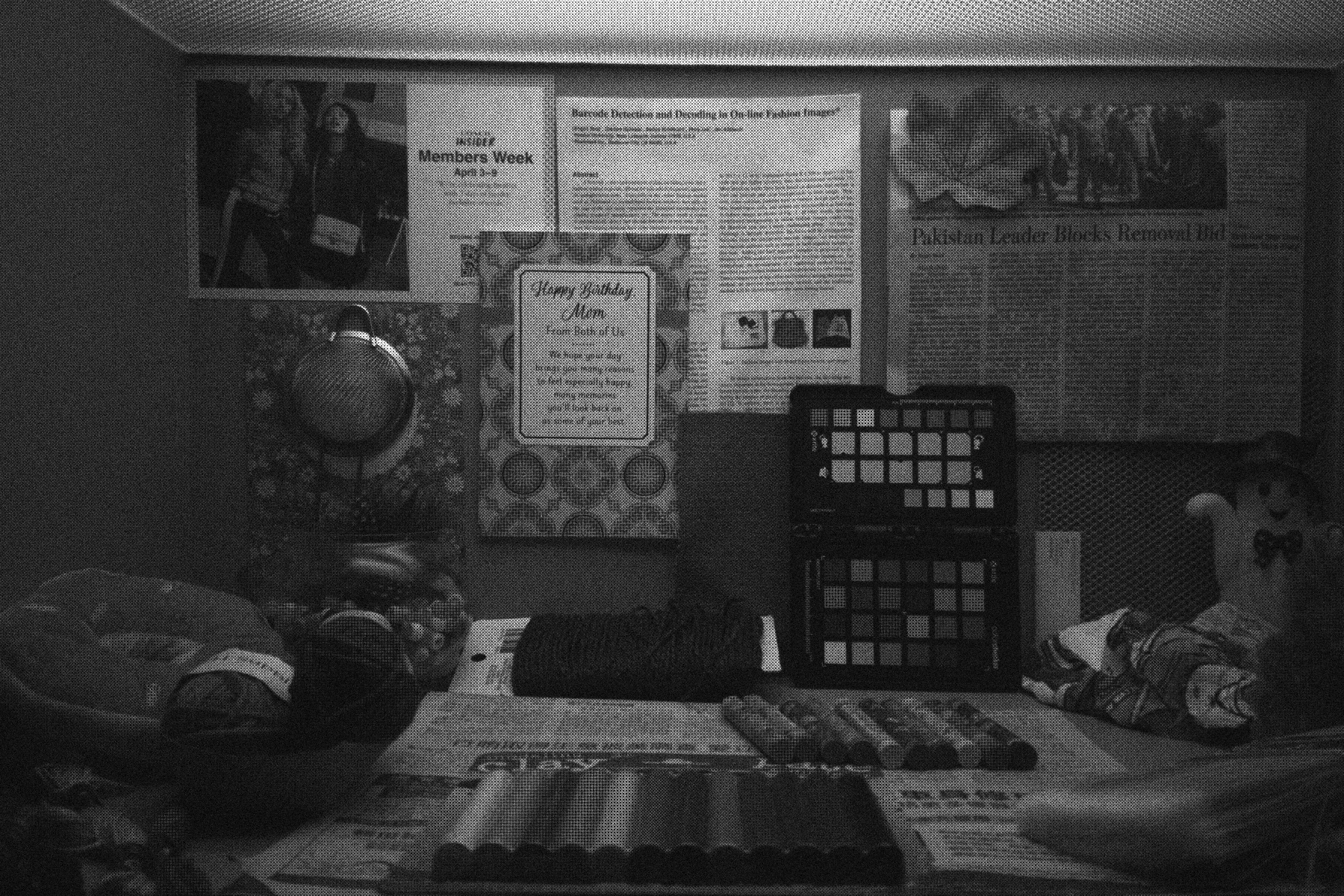}
						\\
						MIPI-Quad-071: Quad Bayer Mosaic 
					\end{tabular}
				\end{adjustbox}
				\begin{adjustbox}{valign=t}
					\begin{tabular}{cccccc}
						\includegraphics[trim={1500 400 200 700  },clip,height=\h \textwidth, width=\w \textwidth]{\name Quad_input/quad_071_fullres_42db_res_mos.png} \hspace{\g} &
						\includegraphics[trim={1500 400 200 700   },clip,height=\h \textwidth, width=\w \textwidth]{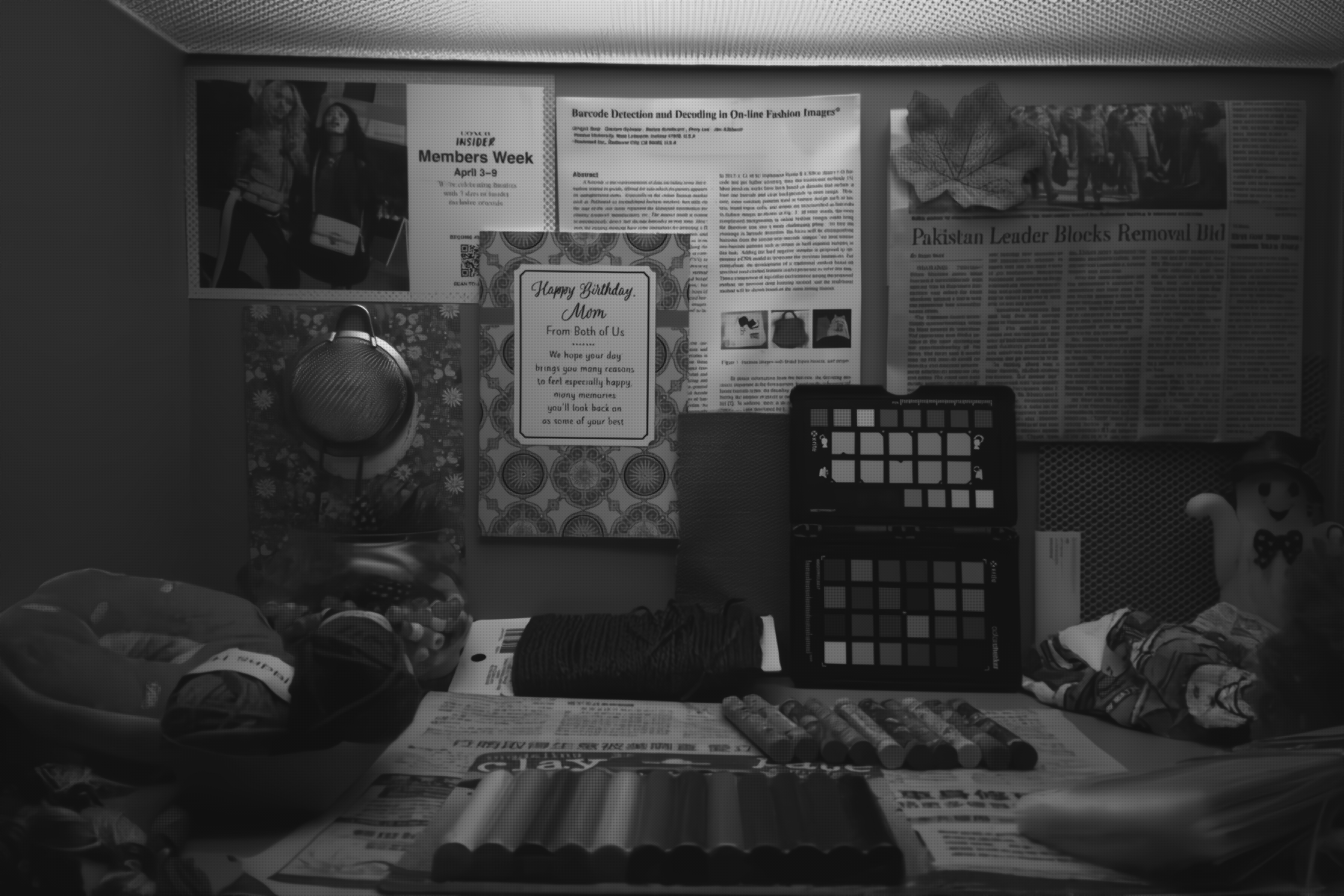} \hspace{\g} &
						\includegraphics[trim={1500 400 200 700   },clip,height=\h \textwidth, width=\w \textwidth]{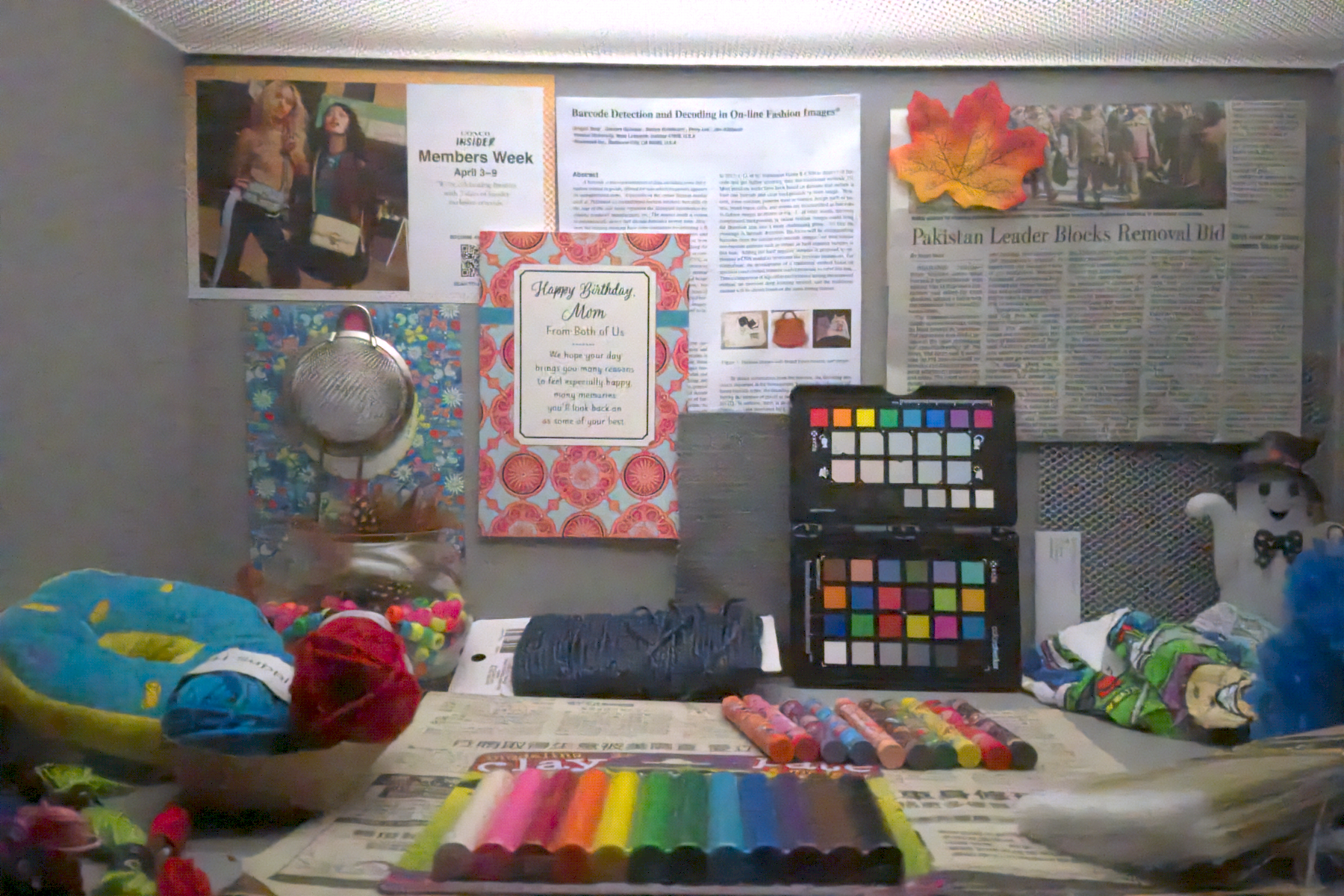} &
						\includegraphics[trim={1500 400 200 700   },clip,height=\h \textwidth, width=\w \textwidth]{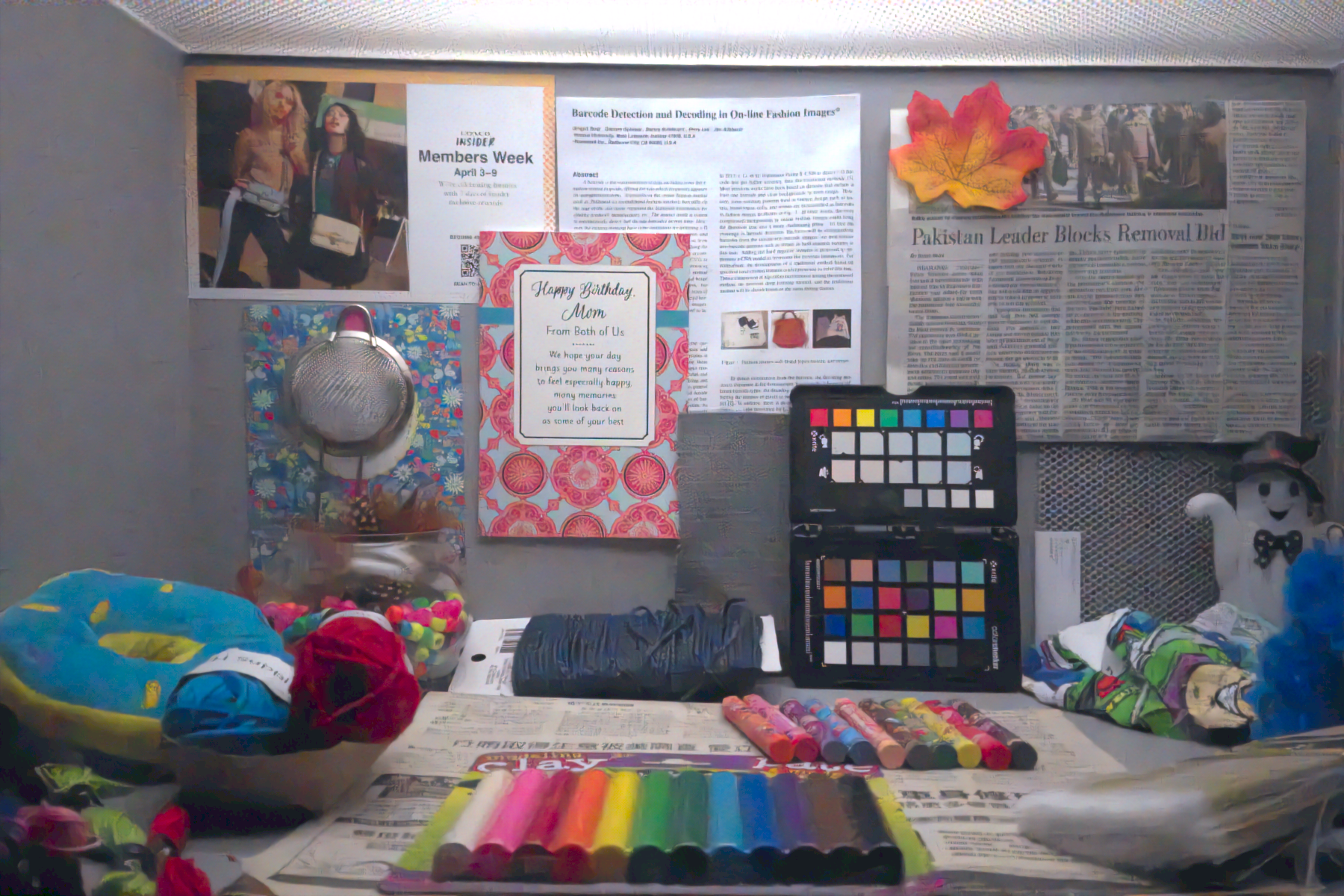} \hspace{\g} 
						\\
						\textbf{Input: Noisy Quad Bayer}  &
						\textbf{Bayer by DJRD (Our)} & MIT~\cite{MIT_deep_joint}&
						PIPNet~\cite{BJDD} 
						\\
            			 & \textbf{KLD = 0.009} &	(28.52, 0.80)
						&
						(28.66, 0.82)
						\\
						\vspace{-2.5mm}
						\\
						\includegraphics[trim={1500 400 200 700   },clip,height=\h \textwidth, width=\w \textwidth]{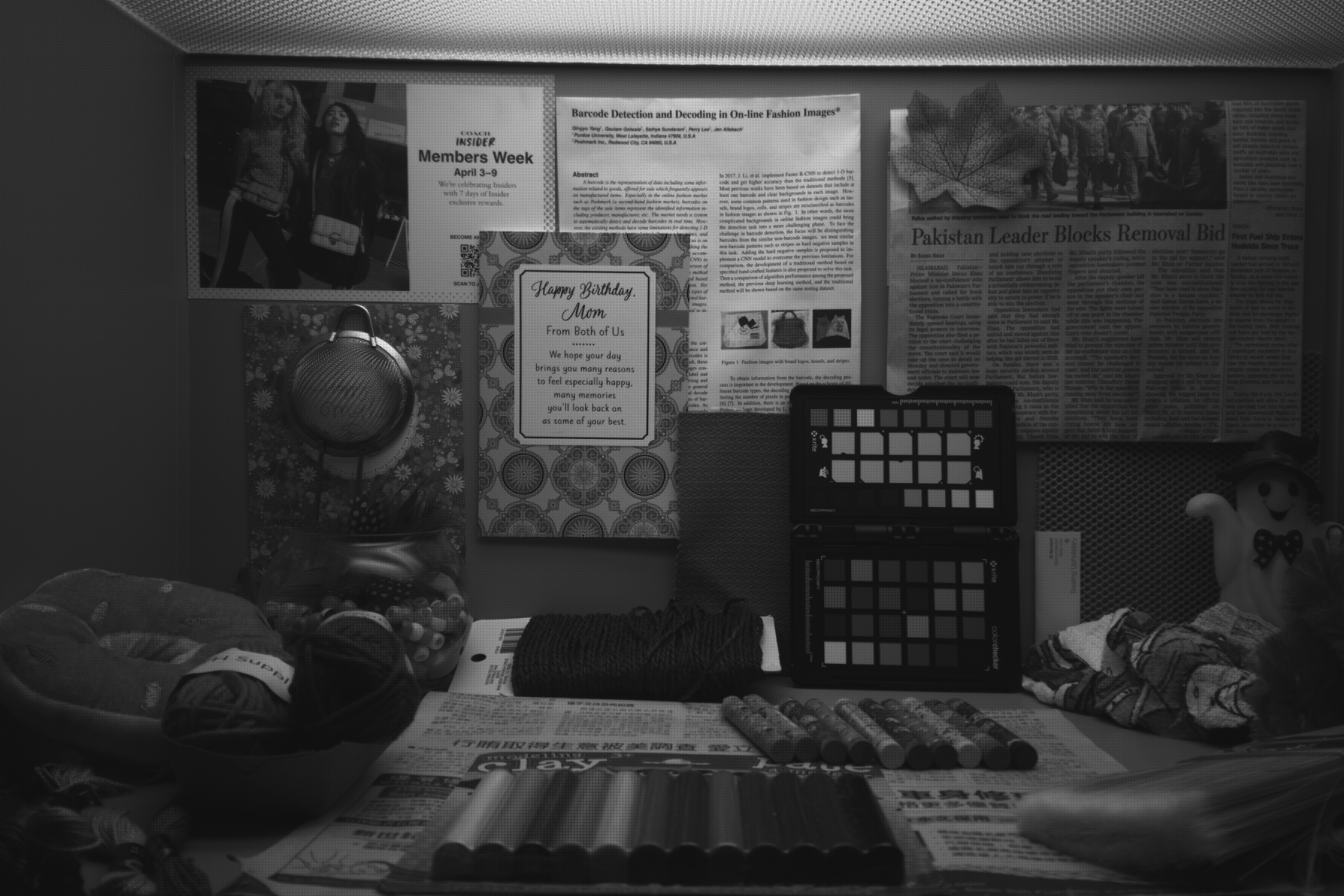} \hspace{\g} &
						\includegraphics[trim={1500 400 200 700  },clip,height=\h \textwidth, width=\w \textwidth]{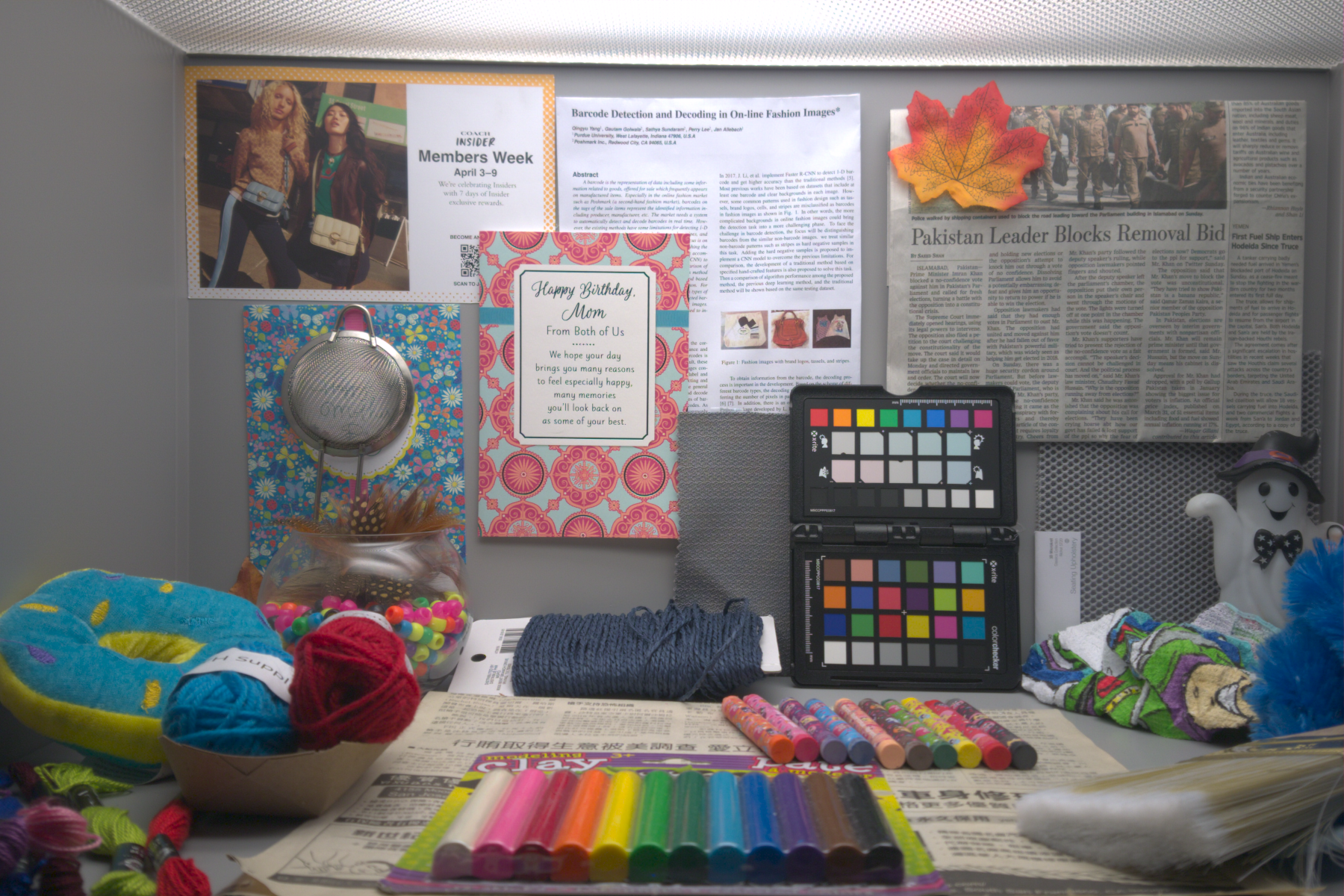} \hspace{\g} &
						\includegraphics[trim={1500 400 200 700   },clip,height=\h \textwidth, width=\w \textwidth]{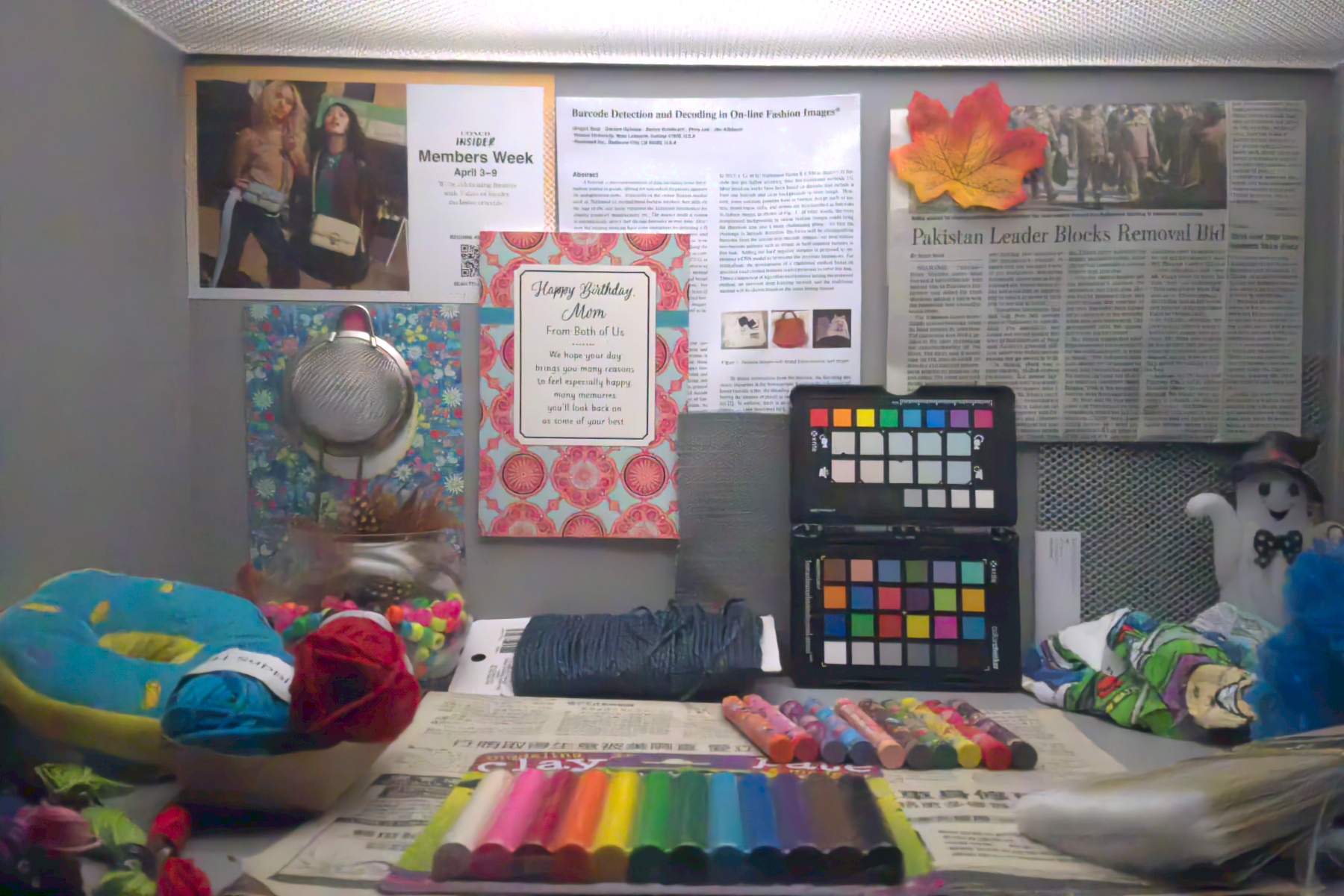}
						\hspace{\g} &		
						\includegraphics[trim={1500 400 200 700   },clip,height=\h \textwidth, width=\w \textwidth]{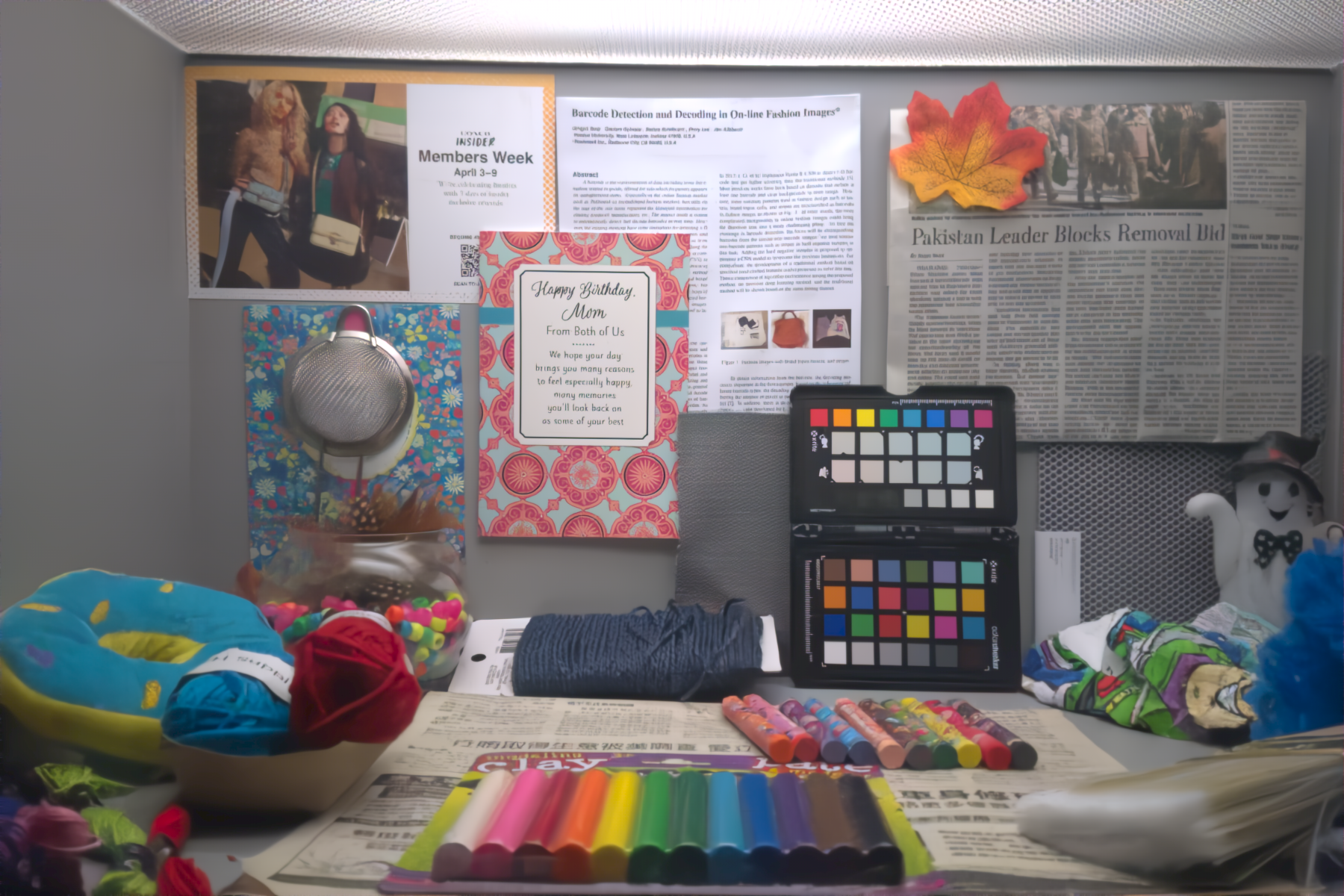} 
						\\ 
						Ground Truth  in Bayer Domain \hspace{\g} & Ground Truth  in RGB Domain  \hspace{\g} &	SAGAN~\cite{SAGAN}
						&
						\textbf{Our} (DJRD+MIT)
						\\
            			  & (PSNR, SSIM) &	(29.85, 0.85)
						&
						\textbf{(31.93, 0.90)}
						\\
					\end{tabular}
				\end{adjustbox}
			\end{tabular}	
		\end{tabular}
	}
% 	\vspace{0.5mm}
% 	\caption{Visual comparison of \textbf{Quad Bayer joint remosaicing and denoising} methods.} %
% 	  \vspace{-2mm}
% 	\label{fig_mipi_071}
% \end{figure*}

% \begin{figure*}[!htbp]
% 	% \captionsetup{font=small}
% 	\centering
% 	\scriptsize
% 	\renewcommand{\h}{0.105}
% 	\renewcommand{\wa}{0.12}
% 	\newcommand{\wb}{0.16}
% 	\renewcommand{\g}{-0.7mm}
% 	\renewcommand{\tabcolsep}{1.8pt}
% 	\renewcommand{\arraystretch}{1}
        \resizebox{0.88\linewidth}{!} {
		\begin{tabular}{cc}			
			\renewcommand{\name}{images/}
			\renewcommand{\h}{0.13}
			\renewcommand{\w}{0.2}
			\begin{tabular}{cc}
				\begin{adjustbox}{valign=t}
					\begin{tabular}{c}%
		         	\includegraphics[trim={625 10 100 70 },clip, width=0.285\textwidth]{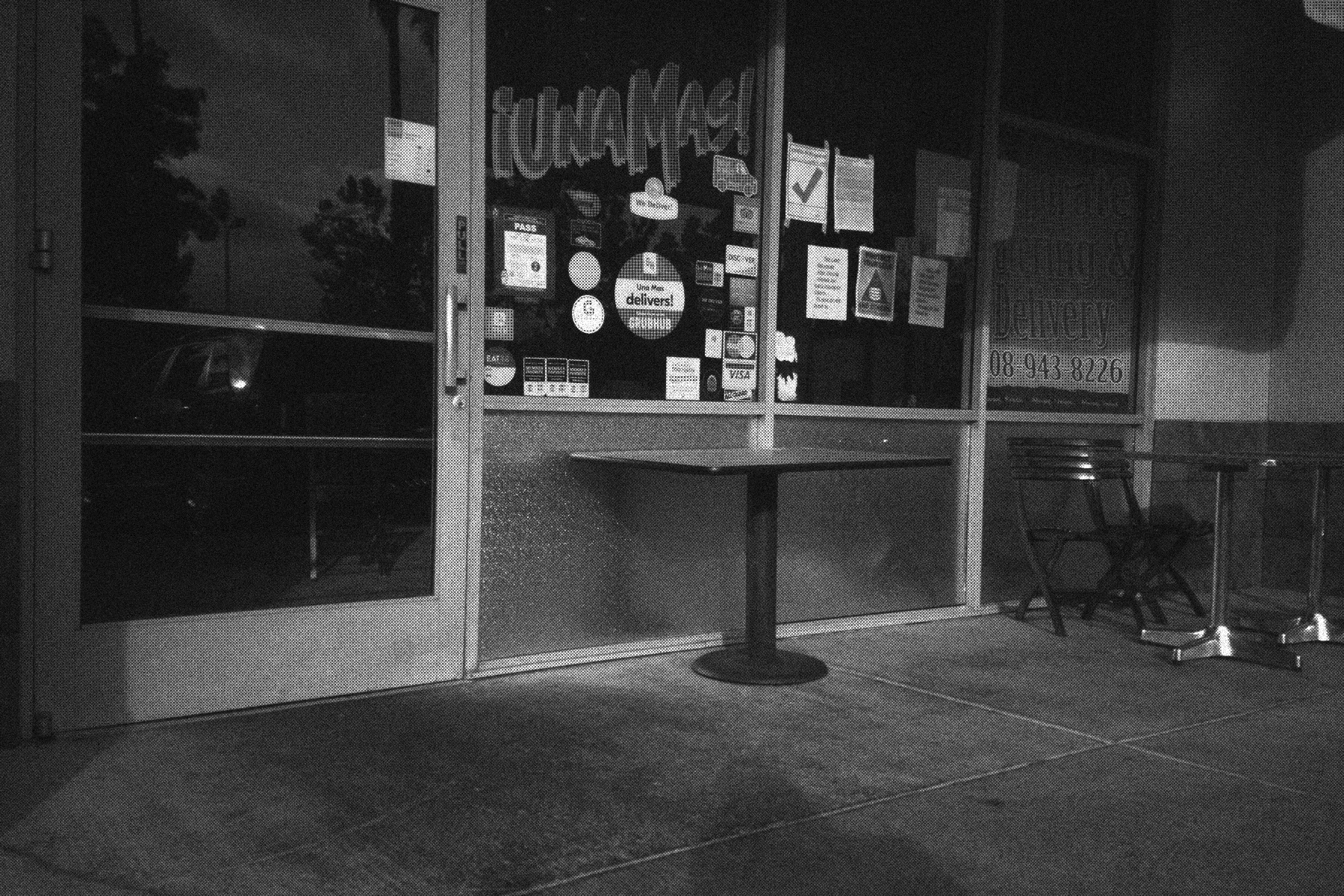}
						\\
						MIPI-Quad-082: Quad Bayer Mosaic 
					\end{tabular}
				\end{adjustbox}
				\begin{adjustbox}{valign=t}
					\begin{tabular}{cccccc}
						\includegraphics[trim={1400 700 300 400 },clip,height=\h \textwidth, width=\w \textwidth]{\name Quad_input/quad_082_fullres_42db_res_mos.png} \hspace{\g} &
						\includegraphics[trim={1400 700 300 400  },clip,height=\h \textwidth, width=\w \textwidth]{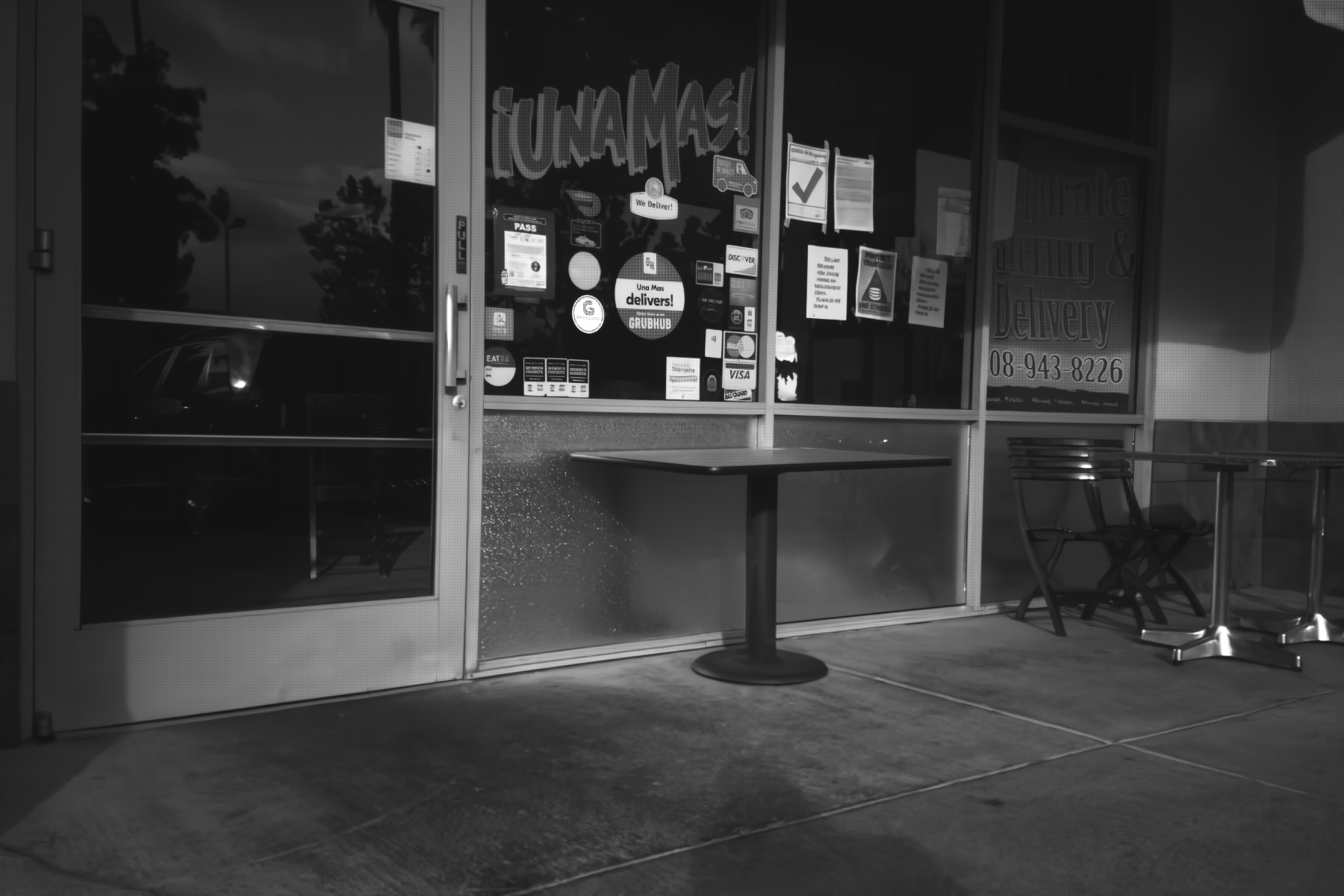} \hspace{\g} &
						\includegraphics[trim={1400 700 300 400  },clip,height=\h \textwidth, width=\w \textwidth]{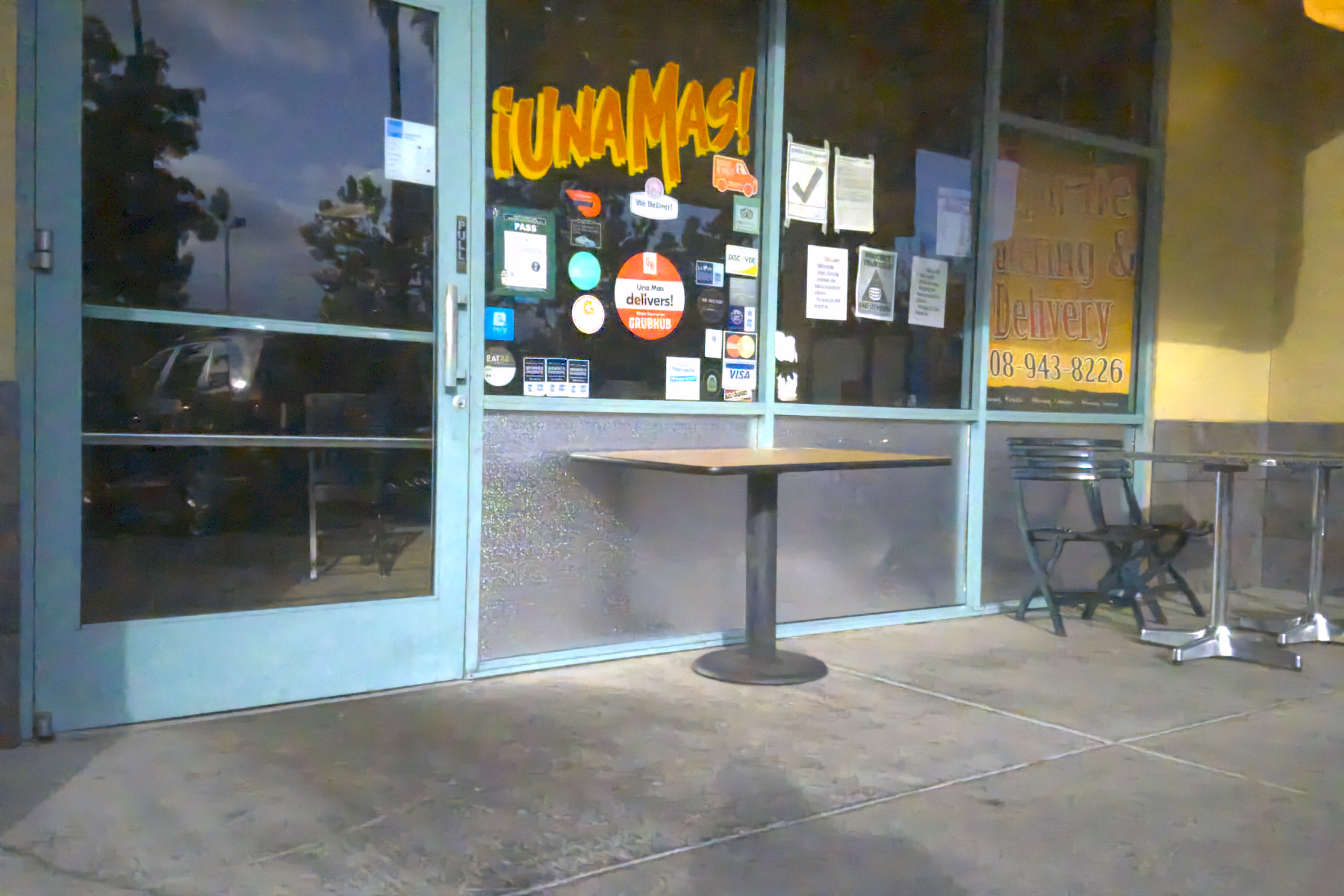} &
						\includegraphics[trim={1400 700 300 400  },clip,height=\h \textwidth, width=\w \textwidth]{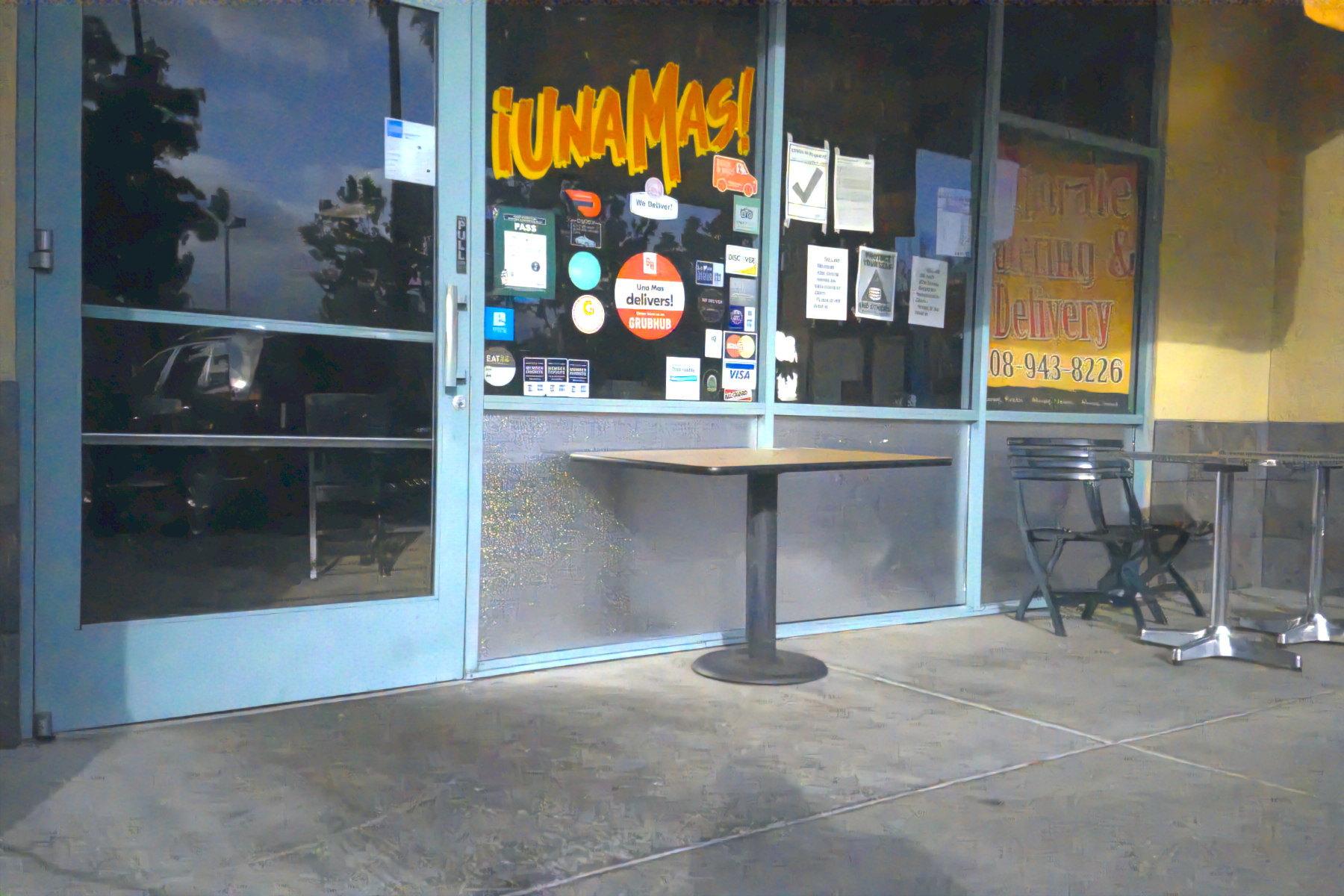} \hspace{\g} 
						\\
						\textbf{Input: Noisy Quad Bayer}  &
						\textbf{Bayer by DJRD(Our)} & MIT~\cite{MIT_deep_joint}&
						PIPNet~\cite{BJDD} 
						\\
                            & \textbf{KLD = 0.012} &	(30.37, 0.81)
						&
						(30.41, 0.81)
						\\
						\vspace{-2.5mm}
						\\
						\includegraphics[trim={1400 700 300 400  },clip,height=\h \textwidth, width=\w \textwidth]{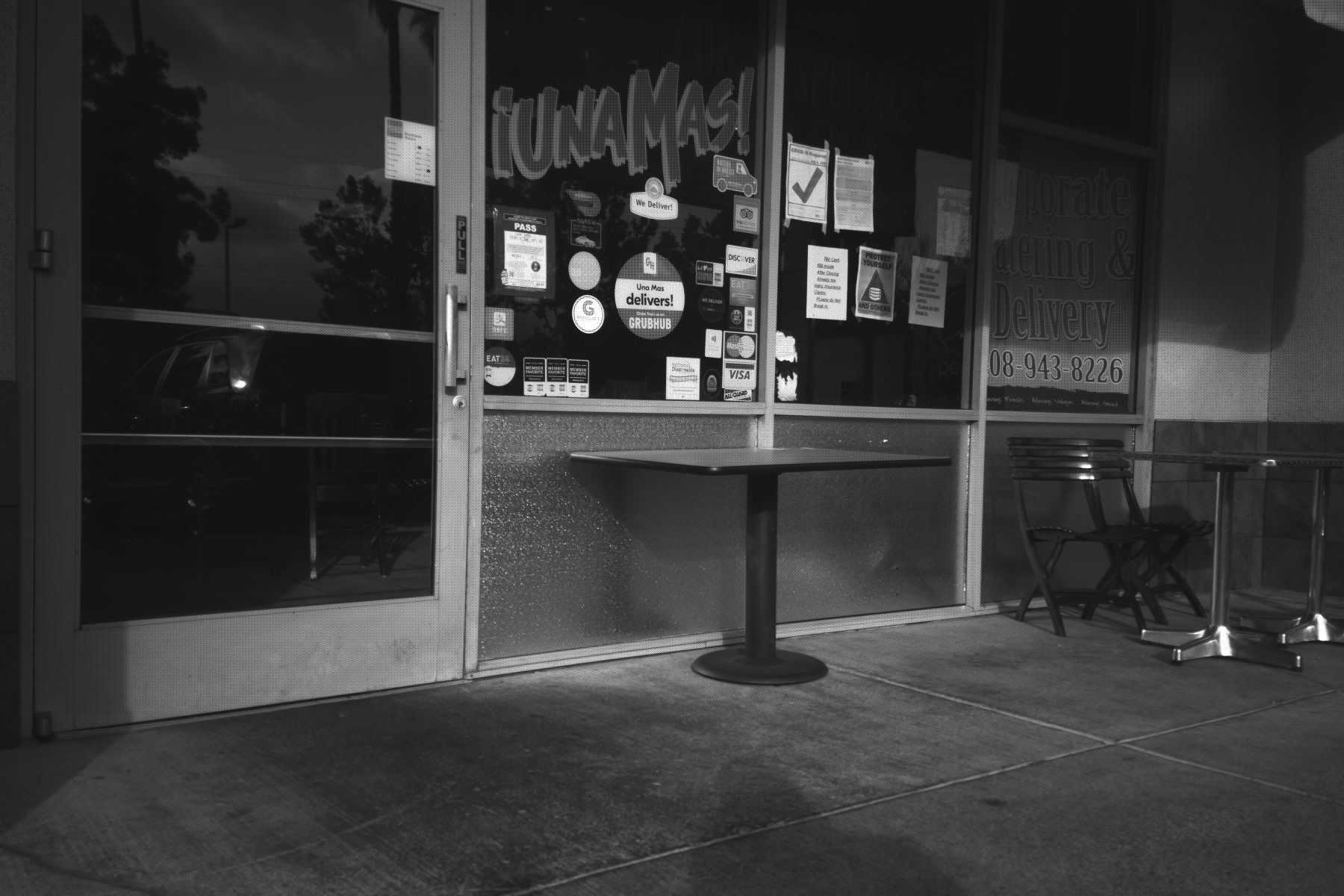} \hspace{\g} &
						\includegraphics[trim={1400 700 300 400  },clip,height=\h \textwidth, width=\w \textwidth]{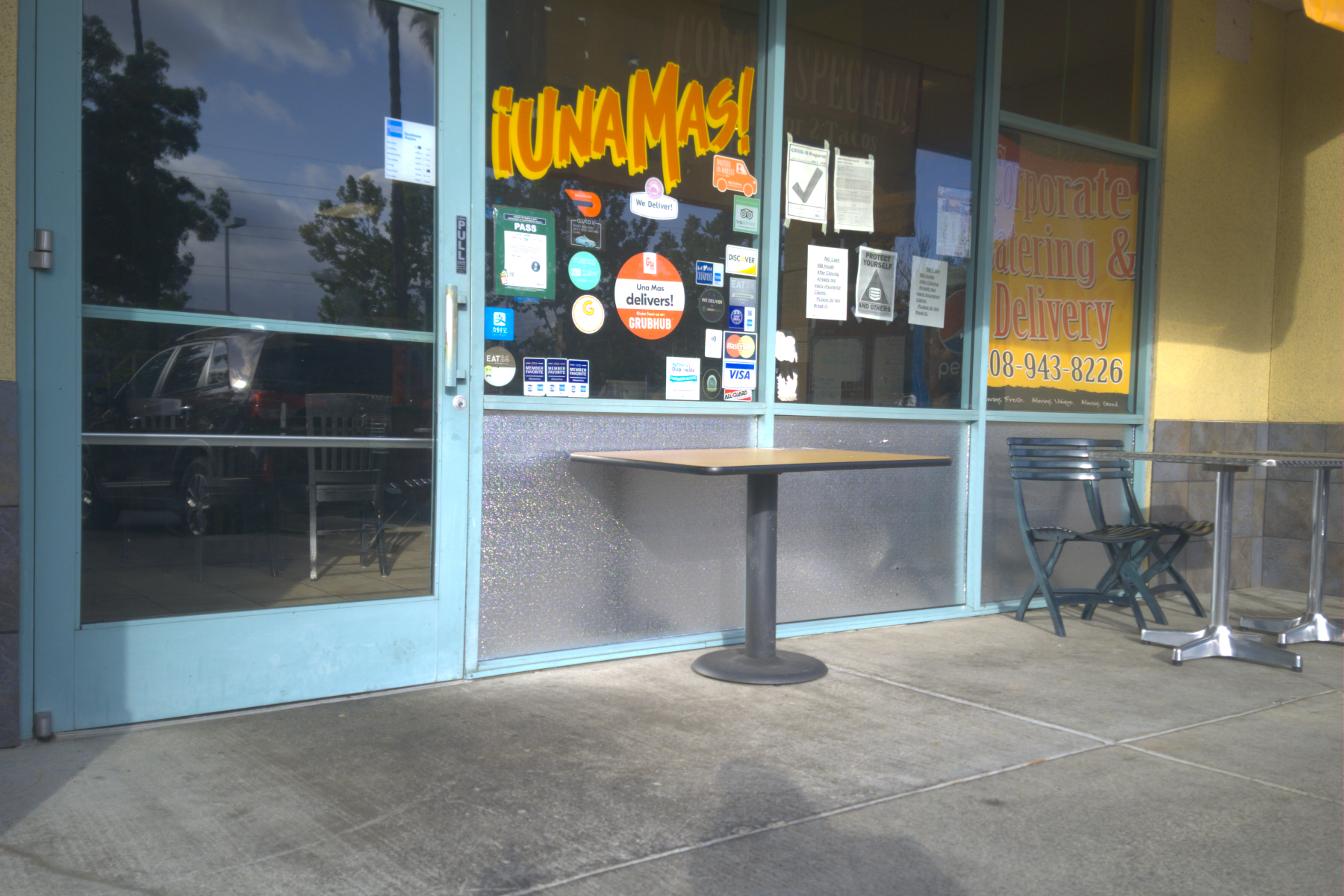} \hspace{\g} &
						\includegraphics[trim={1400 700 300 400   },clip,height=\h \textwidth, width=\w \textwidth]{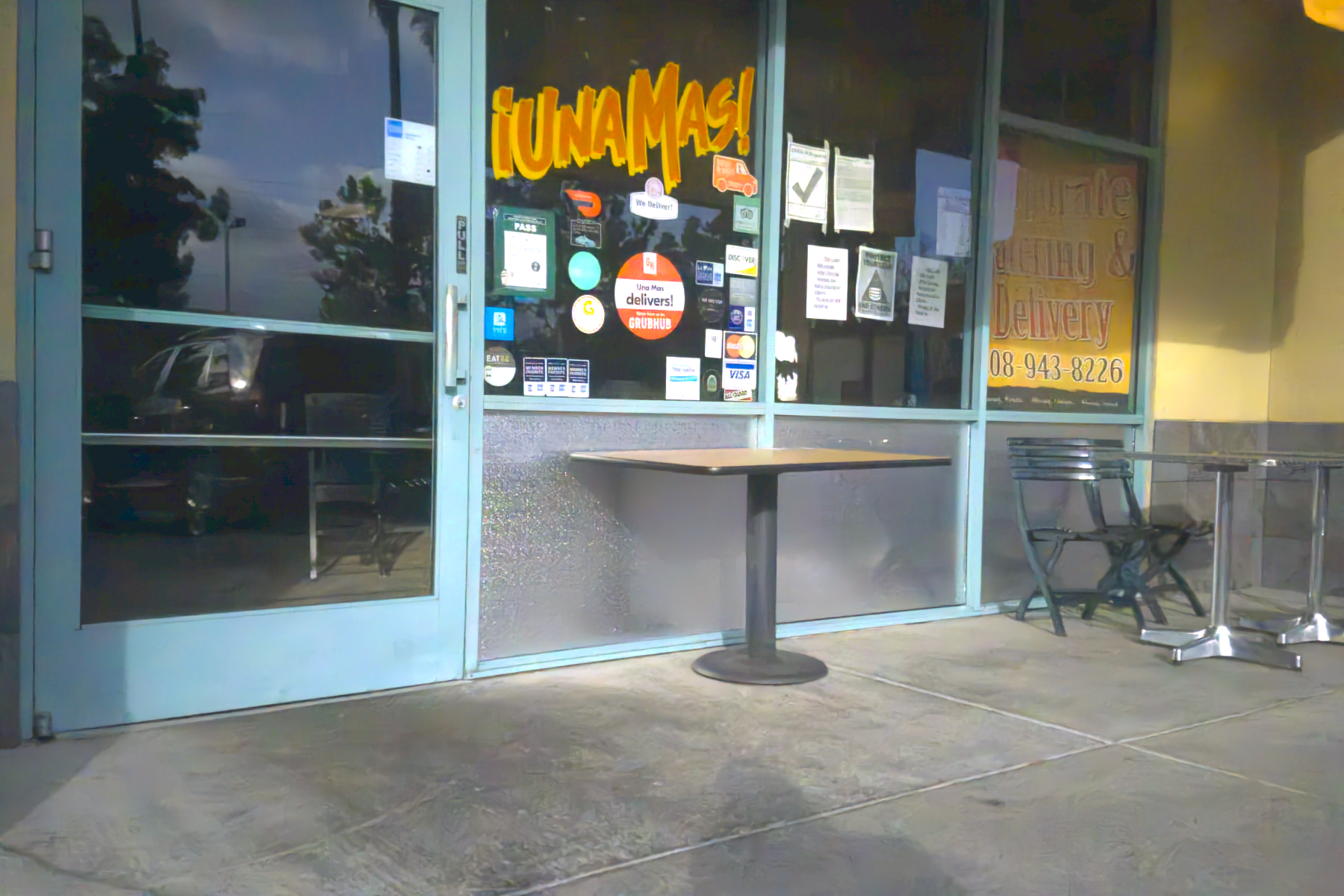}
						\hspace{\g} &		
						\includegraphics[trim={1400 700 300 400  },clip,height=\h \textwidth, width=\w \textwidth]{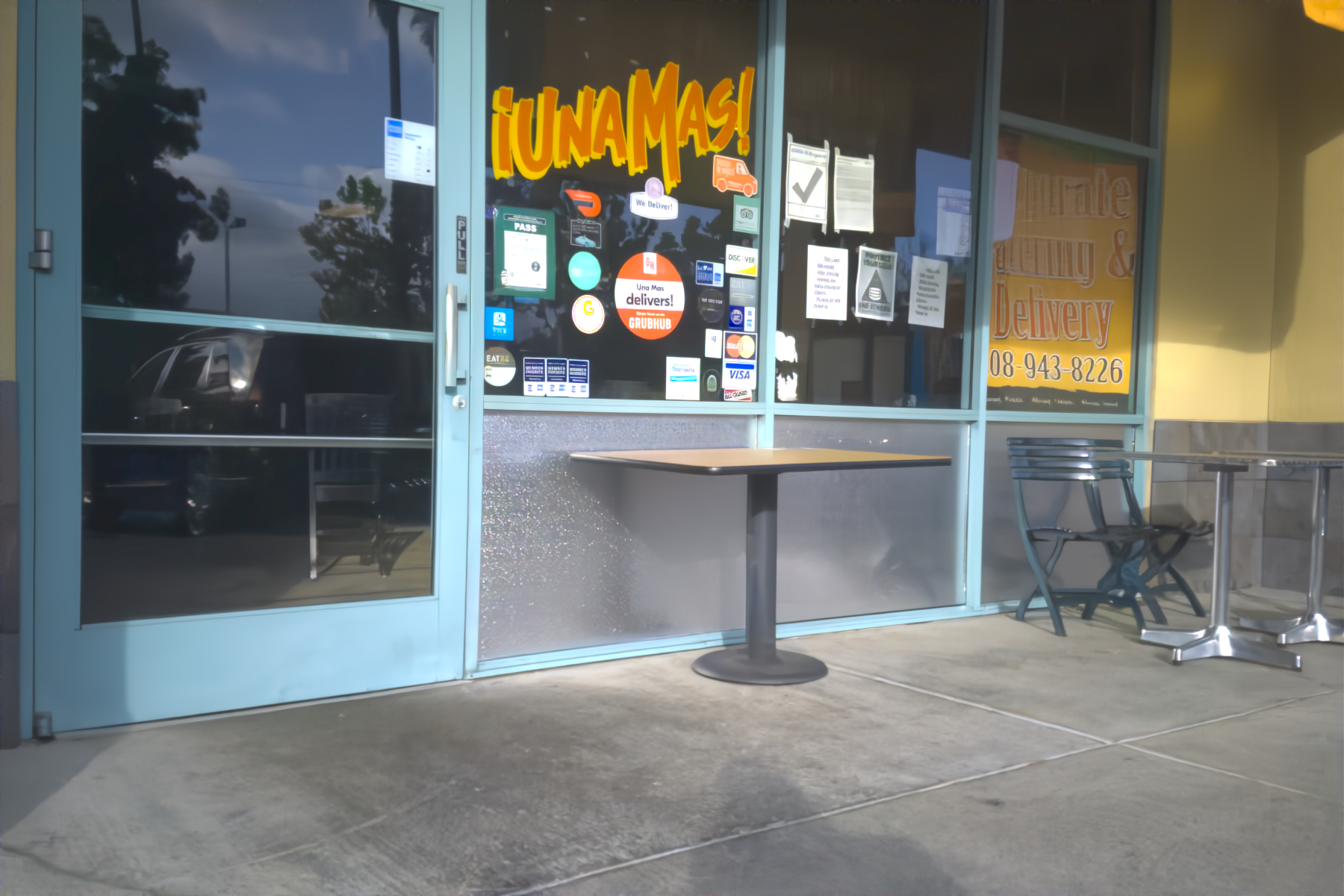} 
						\\ 
						Ground Truth  in Bayer Domain \hspace{\g} & Ground Truth  in RGB Domain  \hspace{\g} &	SAGAN~\cite{SAGAN}
						&
						\textbf{Our} (DJRD+MIT)
						\\
      					& (PSNR, SSIM) &	(31.87, 0.85)
						&
						\textbf{(34.05, 0.90)}
						\\
					\end{tabular}
				\end{adjustbox}
			\end{tabular}	
		\end{tabular}
	}
	\vspace{0mm}
	\caption{Visual comparison of \textbf{Quad Bayer joint remosaicing and denoising} methods om MIPI dataset.} %
	  \vspace{-2mm}
	\label{fig_mipi}
\end{figure*}

\begin{figure*}[!htbp]
	% \captionsetup{font=small}
	\centering
	\scriptsize
	\renewcommand{\h}{0.105}
	\renewcommand{\wa}{0.12}
	\newcommand{\wb}{0.16}
	\renewcommand{\g}{-0.7mm}
	\renewcommand{\tabcolsep}{1.8pt}
	\renewcommand{\arraystretch}{1}
        \resizebox{0.88\linewidth}{!} {
		\begin{tabular}{cc}			
			\renewcommand{\name}{images/urban100/}
			\renewcommand{\h}{0.13}
			\renewcommand{\w}{0.2}
			\begin{tabular}{cc}
				\begin{adjustbox}{valign=t}
					\begin{tabular}{c}%
		         	\includegraphics[trim={0 0 180 0 },clip, width=0.275\textwidth]{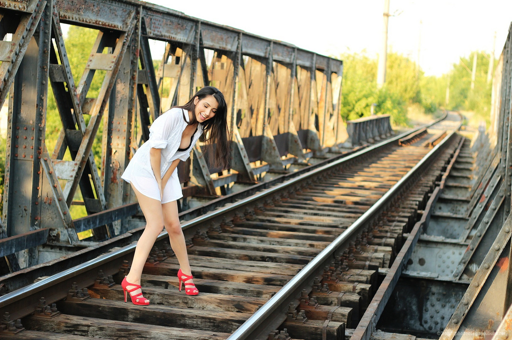}
						\\
						Urban100-007 in RGB Domain
					\end{tabular}
				\end{adjustbox}
				\begin{adjustbox}{valign=t}
					\begin{tabular}{cccccc}
						\includegraphics[trim={170 210 250 80  },clip,height=\h \textwidth, width=\w \textwidth]{\name gt/img_007_SRF_2_HR_0db.png} \hspace{\g} &
						\includegraphics[trim={170 210 250 80  },clip,height=\h \textwidth, width=\w \textwidth]{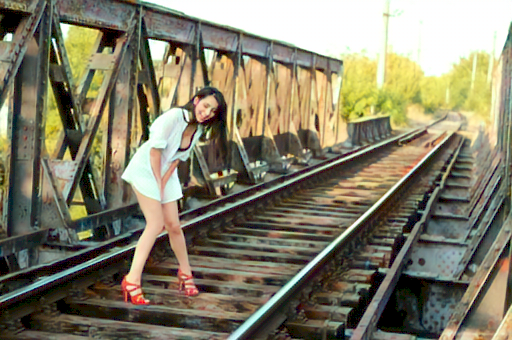} \hspace{\g} &
						\includegraphics[trim={170 210 250 80   },clip,height=\h \textwidth, width=\w \textwidth]{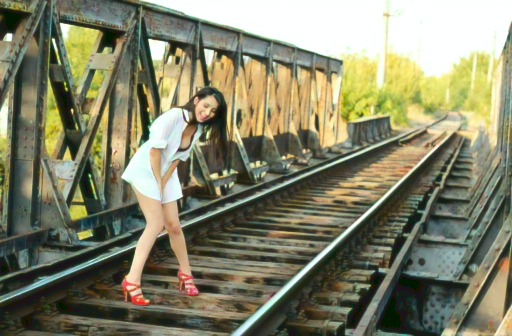}
						\hspace{\g} &		
						\includegraphics[trim={170 210 250 80  },clip,height=\h \textwidth, width=\w \textwidth]{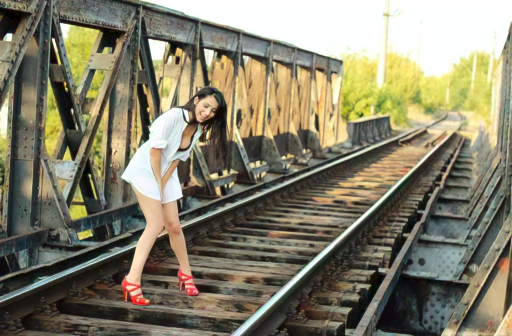} 
						\\
						Ground Truth in RGB Domain \hspace{\g} & MIT~\cite{MIT_deep_joint}  \hspace{\g} &	SAGAN~\cite{SAGAN}
						&
						\textbf{Our} (DJRD+MIT)
						\\
						\vspace{-2.5mm}
						\\
      						\includegraphics[trim={160 20 260 260 },clip,height=\h \textwidth, width=\w \textwidth]{\name gt/img_007_SRF_2_HR_0db.png} \hspace{\g} &
						\includegraphics[trim={160 20 260 260  },clip,height=\h \textwidth, width=\w \textwidth]{\name MIT/img_007_SRF_2_HR_0db.png} \hspace{\g} &
						\includegraphics[trim={160 20 260 260  },clip,height=\h \textwidth, width=\w \textwidth]{\name SAGAN/img_007_SRF_2_HR_0db.png} &
						\includegraphics[trim={160 20 260 260  },clip,height=\h \textwidth, width=\w \textwidth]{\name our/img_007_SRF_2_HR_0db.png} \hspace{\g} 
						\\ 
						Ground Truth in RGB Domain \hspace{\g} & MIT~\cite{MIT_deep_joint}  \hspace{\g} &	SAGAN~\cite{SAGAN}
						&
						\textbf{Our} (DJRD+MIT)
						\\
					\end{tabular}
				\end{adjustbox}
			\end{tabular}	
		\end{tabular}
	}
	\vspace{0mm}
	\caption{Visual comparison of \textbf{Quad Bayer joint remosaicing and denoising} methods on Urban100 dataset.} %
	  \vspace{-2mm}
	\label{fig_urban_07}
\end{figure*}

% To exploit reflectional and rotational invariance in underlying images, we have the same network design trained with the images from the dataset augmented by a possible reflection and three possible rotations.
% The size of the inputs for each training batch is $128 \times 128$, the optimizor we used in training is ADAM.
% % A decreased learning rate is employed to train our model, the initialization is $10^{-4}$, the decreasing rate is $***$ for each $***$ iterations. 
% The platform we used is PyTorch, while the GPU is Titan Xp with 32GB memory.

% \begin{figure}[h]
%     \centering
%     \includegraphics[width=0.48\textwidth]{images/CVPR_comparison_2-1.jpg}
%     \caption{Qualitative evaluation on MIPI challenge dataset with 24dB read noise, shot noise and noise free inputs.}
%     \label{fig: Compare_2}
% \end{figure}

% \begin{figure}
%     \centering
%     \includegraphics[width=0.48\textwidth]{images/CVPR_comparison_5.jpg}
%     \caption{An image of comparison.}
%     \label{fig: Compare_5}
% \end{figure}

% \begin{figure}[h]
%     \centering
%     \includegraphics[width=0.48\textwidth]{images/CVPR_comparison_1-1.jpg}
%     \caption{An image of comparison.}
%     \label{fig: Compare_1}
% \end{figure}

% \begin{figure}
%     \centering
%     \includegraphics[width=0.48\textwidth]{images/CVPR_comparison_1-1_2-1.jpg}
%     \caption{Qualitative evaluation on MIPI dataset with 24dB and 42dB read noise and shot noise.}
%     \label{fig: Compare_6}
% \end{figure}

\subsection{Ablation Studies} \label{ablation_section}

% \emph{Network Analysis:} 
To verify the effectiveness of the employed or designed sub-modules, we decomposed the entire model into degraded models containing only a subset of components and trained them independently. Tab.~\ref{Table:PQI_ablation} presents the PSNR and SSIM values obtained by evaluating the different modules of the proposed network. The results demonstrate that our proposed components significantly enhance the quality of reconstructed images, even in the presence of different levels of noise degradation. More details and discussions are available in the supplementary material.

\begin{table}
\centering
\renewcommand{\arraystretch}{1.3}%
\caption{\textbf{Ablation studies} on six
distinct configurations of our method on MIPI, DN-RM: denoise first then remosaic, DN-RM+hard: fine tune the network with hard cases, Dual+Hard: using the dual-head pipeline fine-tuning on hard cases.}
% and also employing the enhanced reflectional and rotational invariance.}
\label{Table:PQI_ablation}
\scalebox{0.60}{
\begin{tabular}{c|cc|cc|cc|cc}
% \Xcline{1-9}{0.9pt}
Noise Level & \multicolumn{2}{c|}{0 dB} & \multicolumn{2}{c|}{24 dB} & \multicolumn{2}{c|}{42 dB} & \multicolumn{2}{c}{Average}\\
Tactics	&	PSNR	&	SSIM	&	PSNR	&	SSIM	&	PSNR	&	SSIM	&	PSNR	&	SSIM	\\
\Xcline{1-9}{0.4pt}
DN-RM	&	38.35 	&	0.9576 	&	35.51 	&	0.9146 	&	31.91 	&	0.8589 	&	35.26 	&	0.9104 	\\
RM-DN	&	39.21 	&	0.9587 	&	35.57 	&	0.9145 	&	32.01 	&	0.8590 	&	35.59 	&	0.9107 	\\
DN-RM+Hard	&	39.95 	&	0.9635 	&	35.90 	&	0.9170 	&	31.98 	&	0.8594 	&	35.94 	&	0.9133 	\\
RM-DN+Hard	&	40.15 	&	0.9652 	&	35.86 	&	0.9164 	&	32.16 	&	0.8612 	&	36.06 	&	0.9143 	\\
% \underline{PP+Hard}	&	\underline{40.42} 	&	\underline{0.9657} 	&	\underline{36.12} 	&	\underline{0.9186} 	&	\underline{32.34} 	&	\underline{0.8642} 	&	\underline{36.29} 	&	\underline{0.9161} 	\\
% \textbf{PP+Hard+RR}	&	\textbf{40.58} 	&	\textbf{0.9662} 	&	\textbf{36.19} 	&	\textbf{0.9192} 	&	\textbf{32.43} 	&	\textbf{0.8659 }	&	\textbf{36.40} 	&	\textbf{0.9171} 	\\
\textbf{Dual+Hard}	&	\textbf{40.58} 	&	\textbf{0.9662} 	&	\textbf{36.19} 	&	\textbf{0.9192} 	&	\textbf{32.43} 	&	\textbf{0.8659 }	&	\textbf{36.40} 	&	\textbf{0.9171} 	\\
% \Xcline{1-9}{0.9pt}
\end{tabular}}
\vspace{-1mm}
\end{table}

\section{Conclusion}

In this paper, we have presented a dual-head joint remosaicing and denoising network that is capable of converting noisy Quad Bayer and clean classical Bayer mosaic of low-light imaging cameras without any resolution loss. 
Our approach not only facilitates the use of all software and hardware designed for classic Bayer CFA but also allows for any advances in Bayer CFA tools to be directly applicable to Quad sensor. 
Furthermore, our model outperforms the SOTA, yielding a 3dB performance boost under practical noise degradation, which demonstrates the potential of our approach in improving low-light image quality.

{\small
\bibliographystyle{ieee_fullname}
\bibliography{mybibfile}
}

\end{document}